\documentclass{article}
\usepackage{ijcai24}

\usepackage{times}
\usepackage{soul}
\usepackage{url}
\usepackage{hyperref}
\hypersetup{hidelinks}
\usepackage[utf8]{inputenc}
\usepackage[small]{caption}
\usepackage{graphicx}
\usepackage{amsmath}
\usepackage{amsthm}
\usepackage{booktabs}
\usepackage{algorithm}
\usepackage{algorithmic}
\usepackage[switch]{lineno}

\usepackage{caption}
\usepackage{subcaption}
\usepackage{amsmath}
\usepackage{enumitem}
\usepackage{amssymb}

\title{Scaling Law for Language Models Training Considering Batch Size}

\author{
 Xian Shuai$^\dagger$, Yiding Wang$^\dagger$, Yimeng Wu$^\dagger$, Xin Jiang$^\dagger$,  Xiaozhe Ren$^{\dagger}$
 \affiliations
 $^{\dagger}$Huawei Noah’s Ark Lab\\
\emails
\{shuai.xian, wangyiding4, wuyimeng1, Jiang.Xin, renxiaozhe\}@huawei.com
}

\begin{document}

\maketitle

\begin{abstract}
Large language models (LLMs) have made remarkable  advances in recent years, with scaling laws playing a critical role in this rapid progress. 
In this paper, we empirically investigate how a critical hyper-parameter, i.e., the global batch size, influences the LLM training prdocess. 
We begin by training language models ranging from 125 million to 2.6 billion parameters, using up to 300 billion high-quality tokens. Through these experiments, we establish a basic scaling law on model size and training data amount.
We then examine how varying batch sizes and learning rates affect the convergence and generalization of these models. 
Our analysis yields batch size scaling laws under two different cases: with a fixed compute budget, and with a fixed amount of training data. Extrapolation experiments on models of increasing sizes validate our predicted laws, which provides guidance for optimizing LLM training strategies under specific resource constraints. 

\end{abstract}

\section{Introduction}
Recently, LLMs have demonstrated astonishing capabilities in natural language understanding, generation and reasoning. However, training LLMs requires immense compute resources, making the LLM training largely a one-shot, experience-driven endeavor. This contrasts with smaller models, where comprehensive exploration of crucial parameters like batch size and learning rate is feasible. 

To address this challenge, previous works \cite{scalinglawopenai2020,chinchilla} have established scaling laws on small-sized models to guide the cost-effective LLMs training. Nevertheless, they mainly focused on relatively modest batch sizes. As training data and distributed computing systems continue to rapidly expand in scale, there arises a need for increasing the batch size to efficiently utilize the compute resources in parallel and keep high MFU (Model FLOPs Utilization) \cite{megascale,largebs_imagenet}. 
Some studies claim that large batch training can cause the generalization gap and hurt the final performance \cite{largebs_generationgap}. 
Other works observed that batch size has a complicated relationship with the model size, training budget, and the end accuracy \cite{criticalbs2018}. 
In this report, we will systematically explore the impact of batch size on LLMs training. 

First, we build a scaling law benchmark, aiming to validate our experimental platforms. 
In specific, we carefully curate a dataset containing up to 300 billion high-quality tokens, and train GPT-series \cite{gpt3} models from 125M to 2.6B to obtain the basic scaling law on model size $N$ and training dataset amount $D$.
Second, we scale up the batch size to massive levels, up to 32M tokens. This is to explore how such large batch sizes impact training convergence and generalization performance. 
We also recognize that the learning rate (LR) is a closely coupled factor with batch size \cite{largebs_imagenet,donotdecay2018,li2024surge,hoffer2017train}. Therefore, for every batch size, we run three typical LR schemes to investigate the compounded impact. We also further investigate the relationship between the optimal learning rate and the batch size. 

As a result, we extend the fundamental scaling laws by incorporating the batch size as an additional factor. 
Our findings reveal that the optimal batch size can be expressed either as a function of compute budget $C$ when $(N, D)$ lies on the compute-efficient frontier, or as a function of $D$ when $(N, D)$ not necessarily satisfy the compute-efficient frontier. 
To validate these laws, we conduct extrapolation experiments on 4.3B and 7B parameter models, demonstrating the practical effectiveness. Our work presents an investigation of LLMs training scaling laws on the Huawei Ascend infrastructure, and provides detailed guidelines for optimizing LLMs training strategies under different resource constraints. 

\section{BackGround and Related Work}

\subsection{Scaling Law of LLMs Training}
Previous work by Kaplan et al. \cite{scalinglawopenai2020} predicts a pure power law of the cross-entropy loss given the amount of training data and the size of the neural model: 

\begin{equation} \label{eq1}
    L(N, D) = \left[\left(\frac{N_c}{N}\right)^\frac{\alpha_N}{\alpha_D} + \frac{D_c}{D}\right]^{\alpha_D}
\end{equation}

where $L$ is the per-token Cross-Entropy loss, $N$ is the number of non-embedding parameters, $D$ is the amount of training data, and $\alpha_N$, $\alpha_D$, $N_c$, and $D_c$ are constants. 
When either $D$ or $N$ is infinite, this laws degrade to $L_{D=\infty}(N)=(N_c/N)^{\alpha_N}$, and  $L_{N=\infty}(D)=(D_c/D)^{\alpha_D}$. 
By regressing on data across several orders of magnitude in compute, $N$, and $D$, \cite{scalinglawopenai2020} identify a combination of constants for Eq. \ref{eq1}, and suggest that most of the increased compute budget should be allocated to scale up the model size. 
Following this recipe, several LLMs ranging from 175 billion to 530 billion parameters opt to train on around 300 billion tokens \cite{gpt3,mtnlg}.

The “Chinchilla" paper \cite{chinchilla} finds that the scaling laws of \cite{scalinglawopenai2020} is sub-optimal, and improves it by testing at larger scales and with better hyper-parameters \cite{dis_kaplan_chichinlla}. It gives a function form:
\begin{equation}\label{chinchilla_formula}
    L(N, D) = E + \frac{A}{N^{\alpha}} + \frac{B}{D^{\beta}}
\end{equation}
where $E$ denotes the irreducible loss, which can be interpreted as an estimate of the entropy inherent in the underlying data distribution. The other two terms could be transferred into the form of Eq. \ref{eq1} when either $N$ or $D$ approaches infinity. 
We note that the “Chinchilla" law employs a symmetric expression for $N$ and $D$, which omits the $1/D$ expansion \cite{scalinglawopenai2020} and is different from the asymmetric one taken in Eq. \ref{eq1}. 
Based on $\alpha$ and $\beta$ obtained in Eq. \ref{chinchilla_formula}, the “Chinchilla" law suggests that the number of parameters and training tokens should be increased almost equally with more compute. 

Chinchilla's law gives guidance of the optimal allocation of training compute, whereas the inference cost is also important for production-level models. To this end, LLaMA-series models \cite{llama3_herd} take the “over-training" approaches, where the dataset size is much larger than the Chinchilla-optimal.

There exist other scaling laws as well, in areas such as multimodal models \cite{scalinglaw_mm}, transfer learning \cite{scalinglaw_transfer}, mixture-of-experts (MoE) models \cite{scalinglaw_moe}, while they are not the primary focus of this report.

\subsection{Gradient Noise Scale $B_{noise}$} \label{grad_noise_scale}
As mini-batch gradient is just an estimation of the true gradient, larger batches can give a less noisy estimation. There exists a term called the gradient noise scale $B_{noise}$, measuring how large the gradient compared to its variation between different training samples 
\cite{criticalbs2018,scalinglawopenai2020}. 

Specifically, let $\theta\in \mathbb{R}^d$ denote the model parameters, $\eta$ the step size (i.e., learning rate), $G\in \mathbb{R}^d$ the true gradient, and $H\in \mathbb{R}^{d \times d}$ the true Hessian matrix. Then, the loss around parameter $\theta$ under a perturbation $V$ has a quadratic expansion as
\begin{equation}\label{loss_eta_batch}
    L(\theta-\eta V) \approx L(\theta) - \eta G^{T}V + \frac{1}{2}\eta^2 V^THV
\end{equation}
Under the ideal case where the estimated gradient equals the true one $V = G_{est}=G$, by minimizing the loss in Eq. \ref{loss_eta_batch}, we can obtain the maximum learning rate without divergence: 
\begin{equation} \label{max_lr_eq}
    \eta_{max} = \frac{|G|^2}{G^THG}
\end{equation}
However, the estimated gradient $G_{est}$ in reality is noisy due to mini-batch, and it statistically follows:
\begin{equation}
\mathbb{E[G_{est}]}=G ; \ \ cov(G_{est})=\frac{\Sigma}{B}
\end{equation}
where $\Sigma$ is the per-sample co-variance matrix, which varies as a function of $\theta$.
Under such noisy gradient case, minimizing $\mathbb{E}[L(\theta-\eta G_{est})]$ will give the following optimal learning rate and corresponding optimal improvement in loss
\begin{equation}\label{opt_lr}
    \eta_{opt}(B) = \frac{\eta_{max}}{1+B_{noise}/B}; \ \quad \Delta L_{opt}(B) = \frac{\Delta L_{max}}{1+B_{noise}/B}
\end{equation}
where the gradient noise scale $B_{noise} = \frac{tr(H\Sigma)}{G^THG}$ and $\Delta L_{max}$ is a function related to $G$ and $H$.

Aligning with the intuition, the optimal learning rate under the noisy gradient is smaller than the one under the ideal case, i.e., $\eta_{opt}(B) < \eta_{max}$.
In addition, when $B \ll B_{noise}$, increasing the batch size enables a larger learning rate as well as better gradient descent for every iteration step. However, beyond $B_{noise}$, the benefit of further increasing the batch size becomes marginal, thus $B_{noise}$ acts like a turning point for the choosing $B$.

\subsection{Critical Batch Size in LLMs Training}\label{related_cri_bs}
Eq. \ref{opt_lr} indicates that to achieve a loss improvement $\Delta L_{max}$, the full-batch gradient descend only needs a single step, while it needs to take $\delta S = (1+B_{noise}/{B})$ steps when using batch size $B$. By aggregating it over multiple steps of training, there exists a relationship \cite{criticalbs2018,scalinglawopenai2020}
\begin{equation}\label{SE_relation}
    (\frac{S}{S_{min}}-1)(\frac{E}{E_{min}}-1) = \gamma ; \quad \gamma=\frac{(\int\sqrt{B_{noise}}ds)^2}{S_{min}E_{min}}
\end{equation}
where $S$ and $S_{min}$ are the actual and minimum possible steps to reach a specific loss, and $E$ and $E_{min}$ the number of training samples to reach such loss.
Here $\gamma$ represents the amount of variation of the noise scale over the whole training, and lower $\gamma$ denotes the higher training sample utilization efficiency. 
The critical batch size is then defined as:
\begin{equation}
B_{crit} = \frac{E_{min}}{S_{min}} \equiv \frac{\int B_{noise} ds}{\int ds}
\end{equation}
Although $B_{noise}$ may increase as the training goes on, the approximation $B_{crit} \approx B_{noise}$ generally holds. 
Assuming using a fixed batch size $B=B_{crit}$ throughout the training, leveraging Eq. \ref{SE_relation}, we list the combinations of $(E,S,B)$ that should achieve the same training loss theoretically in Table \ref{table_ESB}. 
As shown, $B_{crit}$ provides a good trade-off between $E$ and $S$, and serves as a turning point for using much more steps or using more total data, just like $B_{noise}$.

\begin{table}[h]
\caption{With $S_{min}$, $E_{min}$ and $B_{crit}=\frac{E_{min}}{S_{min}}$, the configuration of $(E, S, B)$ that theoretically should achieve the certain same training loss. }
\label{table_ESB}
\centering
\renewcommand{\arraystretch}{1.6} 
\begin{tabular}{|c|c|c|c|c|c|c|c|}
\hline
$\frac{E}{E_{min}}$ & 1.1 & 1.5 & 2 & 3 & 6 & 11 & 101 \\
\hline
$\frac{S}{S_{min}}$ & 10 & 3 & 2 & 1.5 & 1.2 & 1.1 & 1.01 \\
\hline
$\frac{B}{B_{crit}}$ & 0.1 & 0.5 & 1 & 2 & 5 & 10 & 100 \\
\hline
\end{tabular}
\end{table}

\subsection{Large Batch Size Training with Learning Rate Tuning}

It has been observed empirically that large batch training may suffer from poor generalization ability, as it tends to converge to sharp minima in the loss landscape \cite{largebs_generationgap,lin2020extrapolation}. 
Solutions to this issue include injecting noises \cite{wen2018smoothout}, modifying the Batch Norm function \cite{hoffer2017train}, or the optimizer \cite{lin2020extrapolation}. 
We mainly focus on the learning rate part as it is a critical factor that affects both generalization and optimization of models \cite{lewkowycz2020large}, coupled with the batch size \cite{largebs_imagenet,hoffer2017train,donotdecay2018,li2024surge,minicpm}. 

With distributed synchronous SGD, researchers achieve the training with a large batch size of up to 8,192 images without sacrificing accuracy \cite{largebs_imagenet}, by employing a linear scaling rule for the learning rate with an initial warm-up. 
The interpretation is as follows. Given a series of mini-batches ${\mathcal{B}_j}$ for $0 \leq j <k $ with each of batch-size $B$, and the learning rate $\eta$, the mini-batch Stochastic Gradient Descent after $k$ steps follows a form as:
\begin{equation}
    \theta_{t+k} = \theta_t - \eta \frac{1}{B} \sum_{j<k} \sum_{\mathbf{x} \in \mathcal{B}_j} \nabla l(\mathbf{x}, \theta_{t+j})
\end{equation}
In addition, for training with large batch size $kB$ for a single step, it gives:
\begin{equation}
    \hat{\theta}_{t+1} = \theta_t - \hat{\eta} \frac{1}{kB} \nabla l(\mathbf{x}, \theta_{t}), \mathbf{x} \in \cup_j\mathcal{B}_j  
\end{equation}
If assuming $\nabla l(\mathbf{x}, \theta_{t}) \approx \nabla l(\mathbf{x}, w_{t+j})$ for $j<k$, then the large mini-batch could achieve the similar gradient update with the mini-batch one, i.e., $\hat{\theta}_{t+1} \approx \theta_{t+k}$, as long as enlarging the learning rate by $k$ times that $\hat{\eta} = k\eta$.
This suggests increasing the learning rate linearly with the batch size, which is also consistent with Eq. \ref{opt_lr} when $B\ll B_{noise}$.

On the other hand, besides the SGD-style optimizer, Adam (or its variants) is more widely used in the training of LLMs. 
The update of Adam-style optimizer holds a “sign of gradient” approximation  \cite{li2024surge}
\begin{equation}
\theta_{t+1}=\theta_t-\eta \cdot sign(G_{est})
\end{equation}
By substituting $V=sign(G_{est})$ into Eq. \ref{loss_eta_batch} and minimizing its loss, it gives\footnote{The specific values of $\eta_{max}$ and $B_{noise}$ here are different from those in Sec. \ref{grad_noise_scale}.}:

\begin{equation}\label{lr_opt_adam}
\eta_{opt}(B)=\frac{\eta_{max}}{\frac{1}{2}(\sqrt{\frac{B_{noise}}{B}}+\sqrt{\frac{B}{B_{noise}}})} 
\end{equation}

As a result, the optimal learning rate for Adam-style optimizers will initially increase and then decrease as the batch size grows \cite{li2024surge} for Adam. When $B\ll B_{noise}$, the optimal learning rate grows almost with the square root of $B$, which aligns with the suggestion in previous works \cite{hoffer2017train,lr_func_bs_jmlr}.


\section{Overview}

As introduced in Sec. \cite{scalinglawopenai2020,chinchilla,li2024surge}, four main parameters could largely influence the loss: parameters $N$, data amount $D$, batch size $B$, learning rate $Lr$. 

A summary of the main content is in Table \ref{summary_table}. In Sec. \ref{case1}, similar to those in \cite{scalinglawopenai2020,chinchilla}, we conduct the basic scaling law experiments between parameters amount $N$ and data amount $D$, using GPT-series models of 125M to 350M, 760M, 1.3B and 2.6B parameters, with 300B tokens. 
Subsequently, we explore various combinations of batch sizes and learning rates, while keeping the training tokens with 100B considering the available compute resources. 
We mainly investigate the effect of batch size to answer the following questions:

\begin{enumerate}
\renewcommand{\labelenumi}{\arabic{enumi})}
  \item  With sufficient amount of data, is it feasible to employ large batch sizes for training without causing model generalization issues? (Step-Loss Comparison in  Sec. \ref{case2})
  \item  Considering a fixed compute budget, what is the most compute-efficient batch size to achieve the optimal performance? (FLOPs-Loss Comparison in Sec. \ref{case2})
  \item With fixed amount of data on hand, what is the optimal batch size, or how to allocate the data within a batch and among batches? (Token-Loss Comparison in Sec. \ref{case2} and Sec. \ref{case3})
  \item  The relationship between batch size and the optimal learning rate (Sec. \ref{case4}).
\end{enumerate}



\section{Experiments} \label{exp_4}

\renewcommand{\arraystretch}{1.2}
\begin{table*}[t] 
\centering
\caption{A summary of the content in Sec. \ref{exp_4}. } \label{summary_table}
\vspace{-10pt}
\label{sec:exp:table:other_dynamic_method}
\begin{tabular}{|c|c|c|c|c|c|}
\hline
Subsection  & Parameters & Data & Batch Size &  Learning Rate & Objective  \\ 
\hline
\ref{case1}  & $N$ & $D$  & heuristic  & heuristic   &  $L(N, D) = ?$
, $N_{opt}=?$, $D_{opt}=?$ \\
\ref{case2} & $N$  & $D$  &  B    &   $LR$   & 
With fixed steps/tokens/compute, $B_{opt} = ?$
\\
\ref{case3} & $N$ & fixed  &  B    &   optimal  &  With certain $D$, $B_{opt} = ?$ \\
\ref{case4} & fixed  & fixed &  B    &   $LR$   &  $Loss_{N,D}(B,LR)=?$  \\
\hline
\end{tabular}
\end{table*}

\subsection{The General Law of N and D} \label{case1}

\begin{figure}[t]
    \centering
    \includegraphics[width=\linewidth]{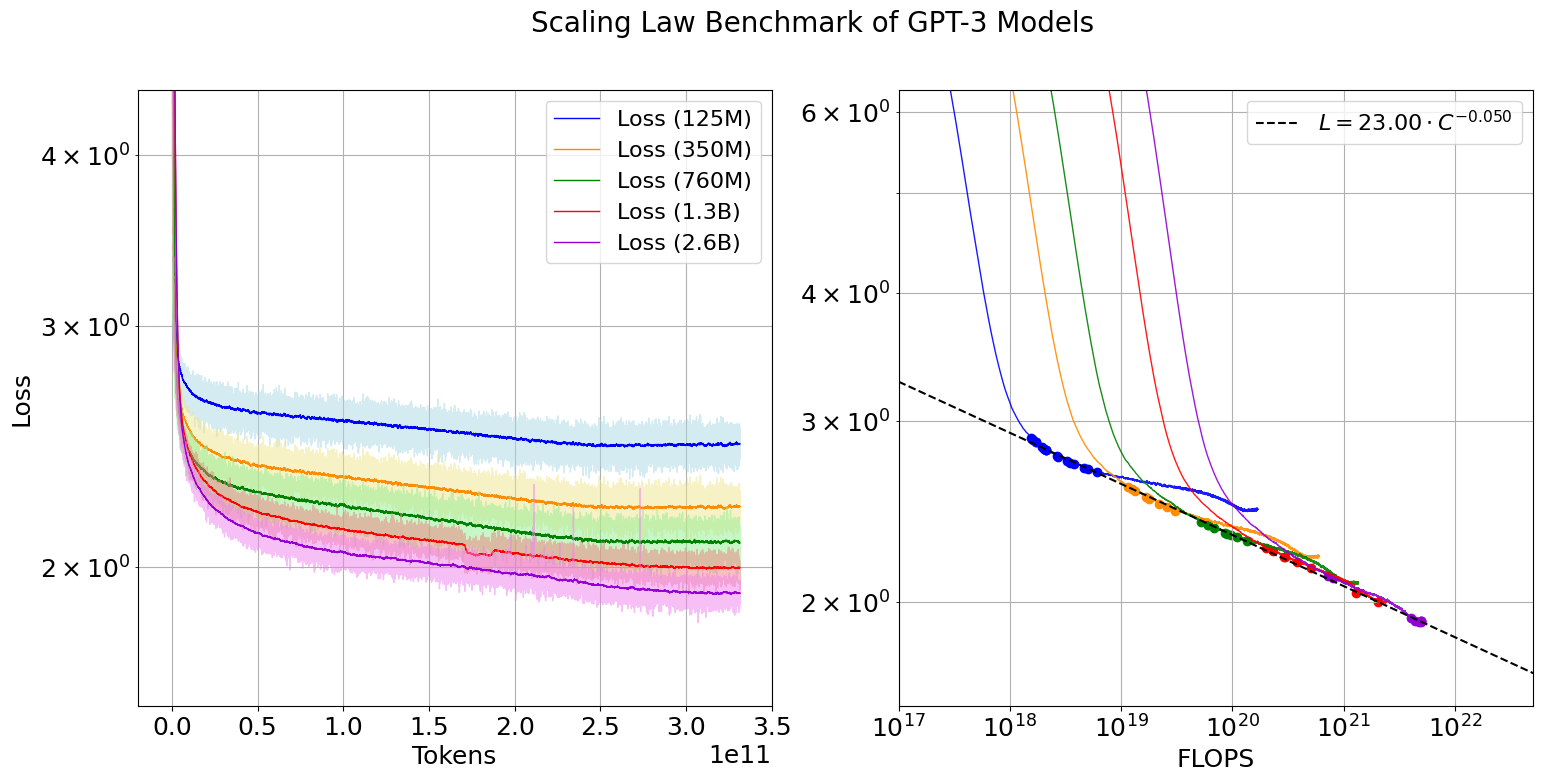}
    \caption{Token-Loss and FLOP-Loss scaling law. }
    \label{law_case1_1}
\end{figure}

\begin{figure}[t]
    \centering
    \includegraphics[width=\linewidth]{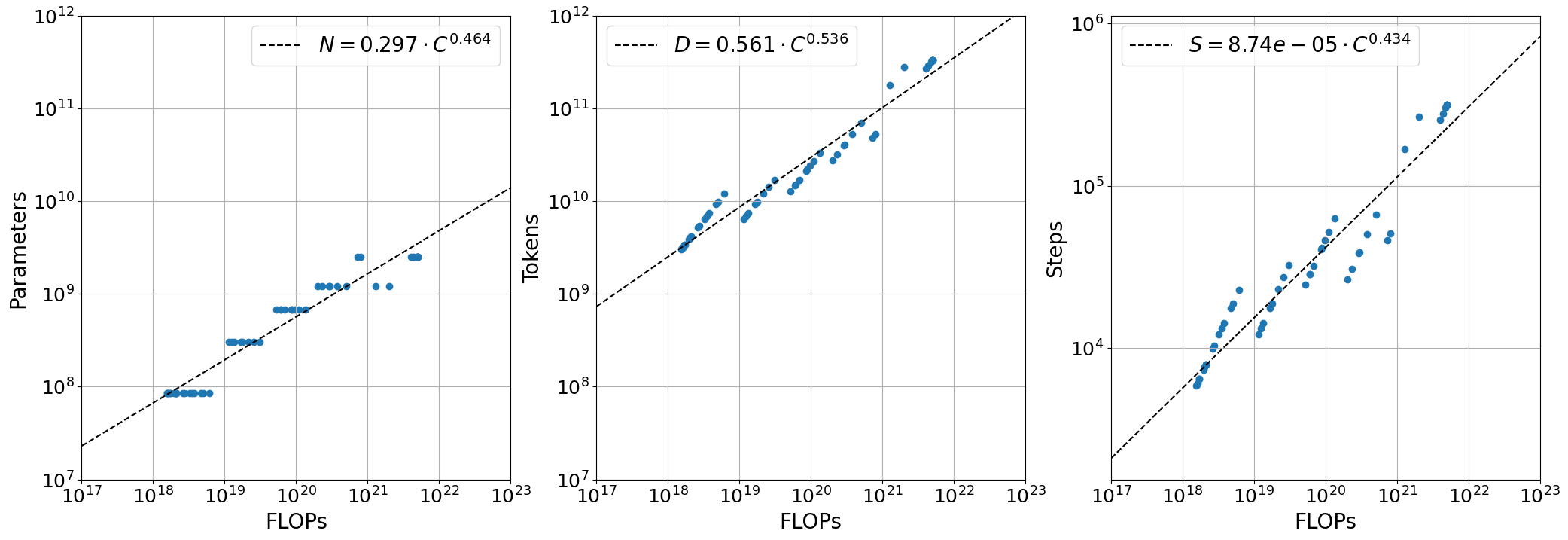}
    \caption{Left and Middle: the optimal number of parameters and training tokens under given FLOPs. Right: The corresponding training steps of the frontier points in Fig. \ref{law_case1_1}.}
    \label{law_case1_2}
\end{figure}

Initially, we adopt the batch size and learning rate setting the same as those used in GPT-3 paper (see Table \ref{gpt3_table}), which have been proven to be heuristically effective. Using these parameters, we plot the token-loss and FLOP-loss relationships using 300B tokens, as shown in Fig. \ref{law_case1_1}. 
We use $FLOPs(N, D) = C \approx 6ND$ for the approximation of compute\footnote{Here, we use the number of non-embedding parameters, and include the last layer of logits head.}. 
The dashed line represents the FLOP-loss frontier across several model sizes, and the dots on each model's curve show the optimal points for that size model regarding the compute-loss effectiveness. The regression on these intersection points gives:
\begin{equation}\label{opt_l_c}
L_{opt} \approx 23.00 \cdot C^{-0.050}
\end{equation}
This demonstrates the minimal loss achievable for a given compute, assuming an adequately large dataset and an appropriately sized model. 
Based on these intersection dots, we further obtain the relationship between FLOP-N and FLOP-D
, as shown in Fig. \ref{law_case1_2}:
\begin{equation}\label{our_L_opt_alpha}
N_{opt} \approx 0.297 \cdot C^{0.464}, \quad  D_{opt} \approx 0.561 \cdot C^{0.536}
\end{equation}

Next, we regress the coefficient in Eq. \ref{chinchilla_formula} using the approach detailed in Appx. \ref{appen_a}, and obtain: 
\begin{equation} \label{our_L_ND}
L(N, D) = 1.48 + \frac{314.35}{N^{0.331}} + \frac{460.51}{D^{0.286}}
\end{equation}
The R-squared value of the regressed loss is 0.962, showing a good estimation of loss (especially for loss after the initial transient period of training).

We compare our results with Keplan's \cite{scalinglawopenai2020}, Chinchilla's laws \cite{chinchilla}, and other industry practices in Table \ref{summary_table_LND}. 
Our law indicates that to achieve the optimal loss as described in Eq. \ref{opt_l_c}, the number of model parameters and tokens should grow proportionally to $C^{0.464}$ and $C^{0.536}$, respectively. This recipe shows a preference for using substantially larger volume of data compared to the one used in GPT-3 or Chinchilla, while aligning closely with the coefficients of Llama-3. Under our law, a 70 billion model should be trained with 3.2e+24 FLOPs and on around 7.7 trillion tokens. 
Our $L(N,D)$  expression
shows lower values in the irreducible loss. 
This may be attributed to our use of different tokenization methods and datasets. 

Notably, using Eq. \ref{our_L_ND} we could also answer a crucial question: to what extent can we compress the model size while maintaining the performance by training with more data?
For example, for a 2.6B model with 1T training token, we can estimate a final loss of around 1.89. However, by expanding the training data to 15T tokens, a much smaller model with only 1B parameters can attain comparable performance, which reduces the inference cost by around 2.5$\times$. This aligns with the current trend of allocating increased training compute in exchange for reducing inference compute. 

\renewcommand{\arraystretch}{1.3}
\begin{table*}[t]
\centering
\caption{Our estimated coefficients and the obtained $L(N,D)$.} \label{summary_table_LND}
\resizebox{\linewidth}{!}{
\begin{tabular}{|c|c|c|c|}
\hline
Approach & Coeff. $\alpha$ for $N_{opt} \propto C^{\alpha}$ & Coeff. $\alpha$ for $D_{opt} \propto C^{\alpha}$ & $L(N,D)$  \\
\hline
GPT-3 \cite{scalinglawopenai2020} & 0.73 & 0.27 & $L(N, D)=[(\frac{8.8 \times 10^{13}}{N})^{0.8} + \frac{5.4\times 10^{13}}{D}]^{0.095}$  \\
Chinchilla \cite{chinchilla} & 0.49 & 0.51 & $L(N, D) = 1.69 + \frac{406.4}{N^{0.34}} + \frac{410.7}{D^{0.28}}$  \\
PaLM-2 \cite{palm2} & 0.49 & 0.51 & - \\
DeepSeek-LLM \cite{deepseek-llm} & 0.524 & 0.476 & - \\

Llama-3 \cite{llama3_herd} & 0.47 & 0.53 & - \\
Ours & 0.464 & 0.536 & $L(N, D) = 1.48 + \frac{314.35}{N^{0.331}} + \frac{460.51}{D^{0.286}}$ \\
\hline
\end{tabular}
}
\end{table*}

\subsection{Comparison under Varied Batch Sizes and Learning Rates} \label{case2}


In this part, we explore the optimal batch size under three different cases: with a fixed number of training steps, with a fixed number of tokens, or with a pre-defined limit of training FLOPs.

\begin{figure}[t]
    \centering
    \includegraphics[width=\linewidth]{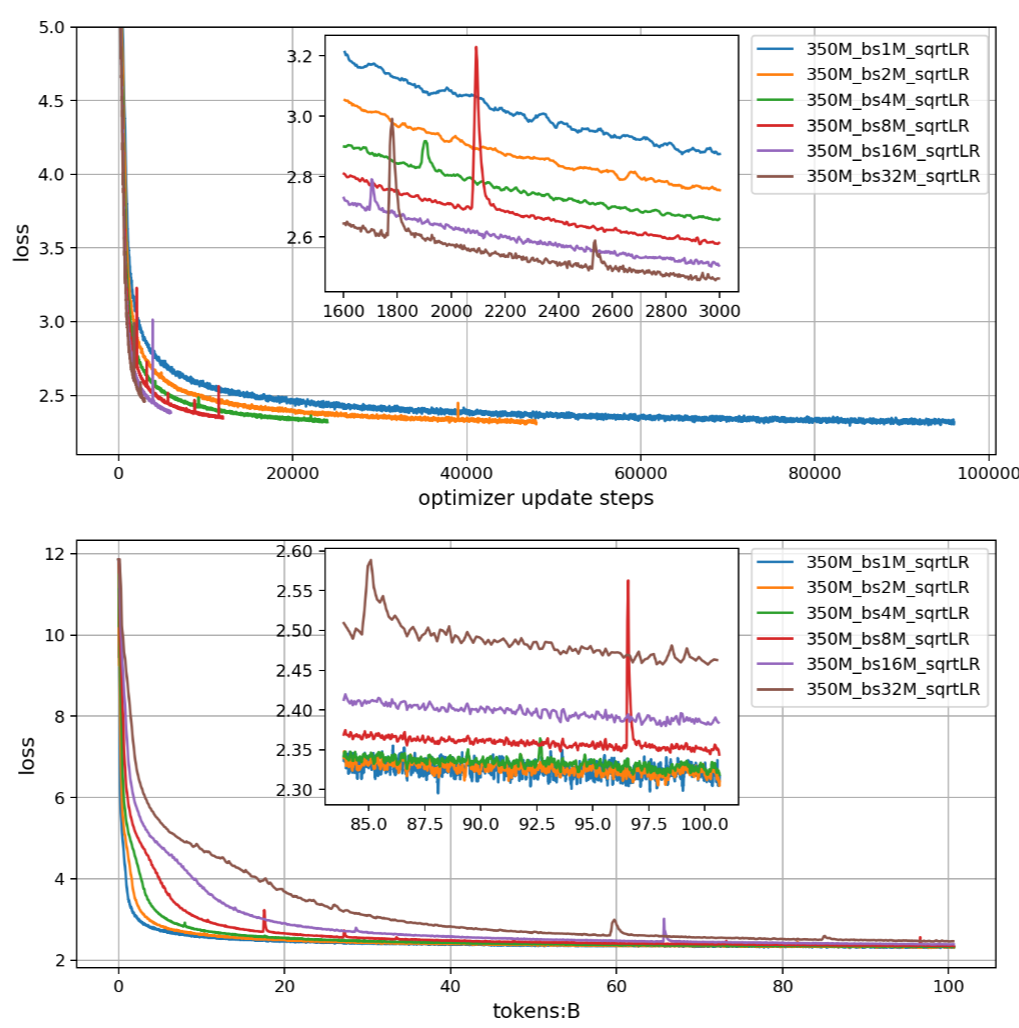}
    \caption{The \textit{upper} and \textit{lower} plots present the loss-step and loss-token curves of the 350M model, respectively. The global batch sizes range from 1M to 32M tokens, with the learning rate scaling with the square root of the batch size. As all experiments use 100B tokens, larger batch sizes result in fewer training steps. More results in Appx. \ref{appen_b}. }
    \label{law_case2_1}
\end{figure}

\subsubsection{Step-Loss Comparison}
A typical Step-Loss plot of different batch sizes is shown in Fig. \ref{law_case2_1}. As illustrated, large batch sizes achieve much lower per-step training loss compared to smaller batch sizes. Additional results in Appx. \ref{appen_b} further show that this phenomenon holds across almost all model sizes, ranging from 350M to 2.6B parameters. This observation aligns with the theoretic intuition in Sec. \ref{grad_noise_scale} that the larger batch size reduces the batch noise and is beneficial for model training. 
Consequently, we anticipate that the large batch size itself is not a main cause of optimization challenges, and when the data amount is large enough to support sufficient training steps, using a substantially larger batch size for training is better than using a smaller one.

\subsubsection{Token-Loss Comparison}

However, larger batch training exhibits limitations in the Token-Loss perspective, primarily due to the reduced number of iteration steps. 
As shown, when employing a batch size as large as 32M tokens per iteration, a dataset of 100B tokens will only support around 3,000 steps of training. 
A workaround to compensate for the iteration step reduction is increasing the learning rate. Typically, there are two learning rate enlarging schemes: 

\begin{enumerate}
\renewcommand{\labelenumi}{\arabic{enumi})}
  \item Linear increasing with batch size.
  \item Square Root increasing with batch size.
\end{enumerate}
From the lower panel of Fig. \ref{law_case2_1} (with more results in Appx. \ref{appen_b}.), we observe that under appropriate learning rate adjustments, larger batch sizes (e.g., 4M) could achieve comparable performance with the smaller ones (e.g., 1M), even in Token-Loss metric. 
Here, we only qualitatively show the potential of using large batch size for training under given tokens, while a detailed quantitative analysis is left in Sec. \ref{case3}.


\subsubsection{FLOPs-Loss Comparison}\label{flop-loss-opt}
\begin{figure}[t]
    \centering
    \includegraphics[width=1.\linewidth]{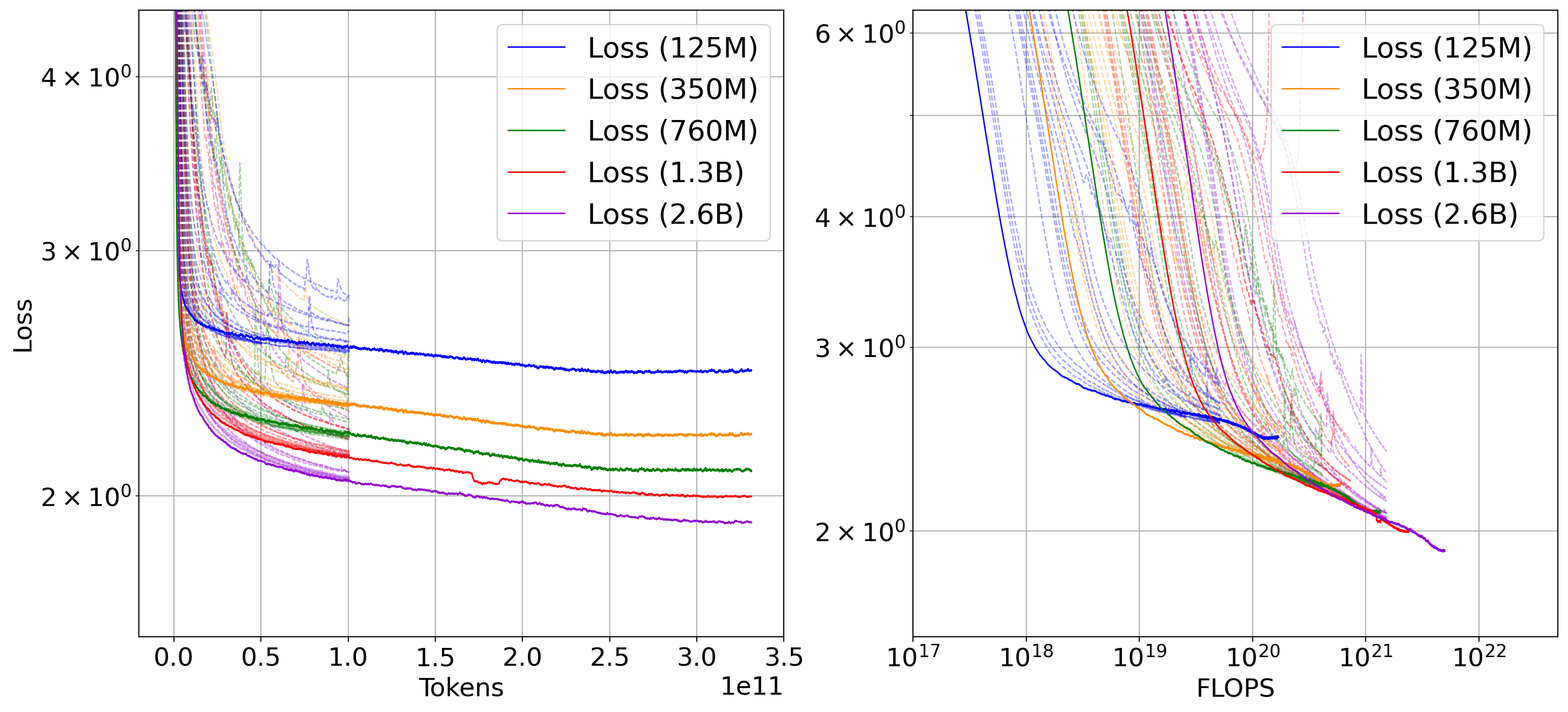}
    \caption{The solid curves represent the curves depicted in Fig. \ref{law_case1_1},  trained with 300B tokens. The dashed, fading curves illustrate the loss using different batch sizes and learning rates, running with 100B tokens. Zoom in for better viewing. }
    \label{law_fig_case2_3}
\end{figure}

In Fig. \ref{law_fig_case2_3}, we overlay additional curves with different batch sizes ranging from 1M to 32M, along with the previously mentioned three learning rate schemes, on top of the curves already depicted in Fig. \ref{law_case1_1}. 
As shown in the left panel, a few dotted curves fall below the corresponding solid curves with the same model size at 100B tokens, indicating higher token-loss efficiency. 
However, this does not alter the FLOP-loss frontier in the right figure. Therefore, we can still utilize the regression results in Fig. \ref{law_case1_2}. \
As the amount of data is also the product of the training steps and the batch size, we have:
\begin{equation}\label{our_opt_bs}
S_{opt} \approx 8.74 \times 10^{-5} \cdot C^{0.434}, \quad  B_{opt} \approx 6.42 \times 10^{3} \cdot C^{0.102}
\end{equation}

This indicates that given a fixed compute budget, to achieve the lowest loss, the compute should be allocated with $N_{opt} \propto C^{0.464}, S_{opt} \propto C^{0.434}, B_{opt} \propto C^{0.102}$. 
With more compute budget, the optimal number of training tokens and batch size both increase. 

Notably, the FLOP-Loss frontier in Fig. \ref{law_fig_case2_3} is obtained using batch sizes equal to or larger than 0.5M tokens. Therefore, we assume Eq. \ref{our_opt_bs} primarily holds for $B_{opt}>0.5M$ and $C>(\frac{0.5M}{6.42 \times 10^{3}})^{\frac{1}{0.102}} \approx 5 \times 10^{18}$ FLOPs. This is because smaller batch sizes could result in a new FLOP-loss frontier, changing the regression in Fig. \ref{law_case1_2}.


\subsection{The Law between D and B with Optimal LR} \label{case3}

\begin{figure*}[t]
    \centering
    \includegraphics[width=\linewidth]{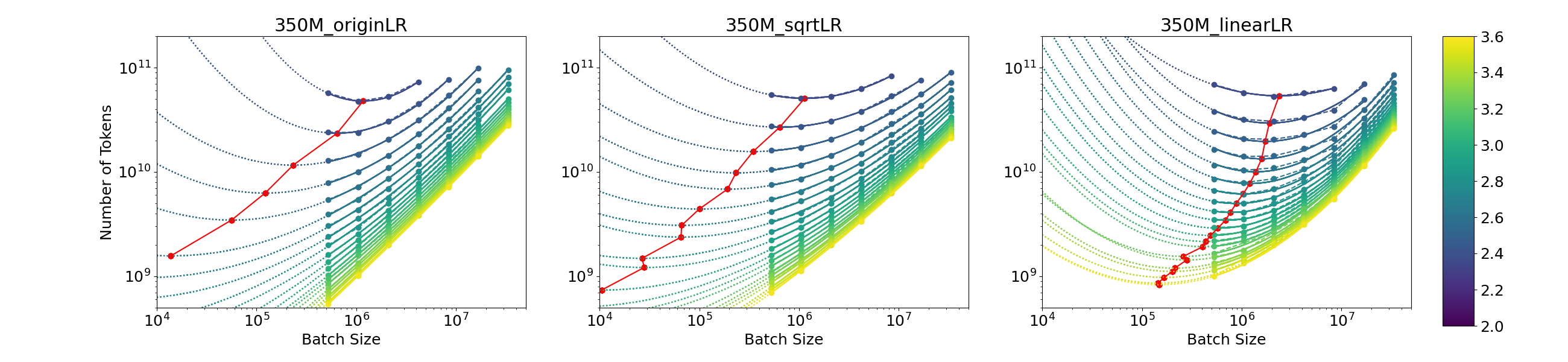}
    \caption{Loss contours of 350M model with different batch sizes and training data amount. Lighter colors denote higher loss. The dotted segments indicate areas that are not empirically obtained but rather fitted. Red points are the lowest point of the parabolas of each loss contour, showing the trend of optimal batch size across the amount of training data. More results in Fig. \ref{law_case3_1_all}. }
    \label{law_case3_1}
\end{figure*}

\begin{figure}[t]
    \centering
    \includegraphics[width=1.\linewidth]{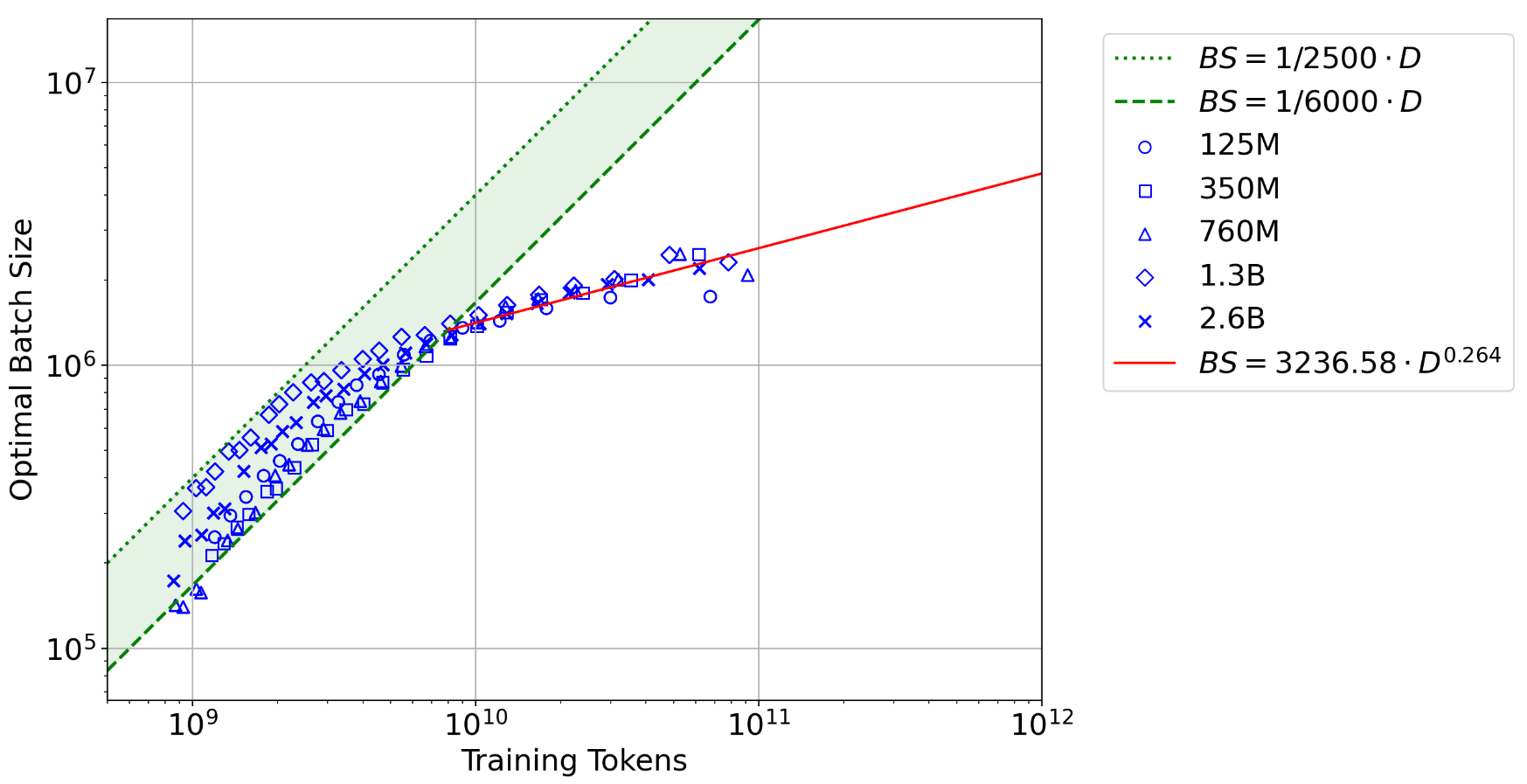}
    \caption{The optimal batch sizes against the available amount of training tokens. The data points mainly fall into two regions: within the green area and outside of it.  }
    \label{law_fig_case3_1_2}
\end{figure}


In Sec. \ref{flop-loss-opt}, we have already shown the optimal selection of $N$, $D$, and $B$ for a given FLOP budget. 
However, a more practical scenario is that $N$ and $D$ are pre-determined and do not conform to the optimal relationship described in Eq. \ref{our_L_opt_alpha}. 
Therefore, we consider with only a fixed amount of training tokens on hand (e.g., 100B), what batch size should be set for different model size $N$. 
In Fig. \ref{law_case3_1} (with more results in Fig. \ref{law_case3_1_all}), we plot the loss contour map of 350M model with different batch size and data amount. We observe:
\begin{enumerate}
\renewcommand{\labelenumi}{\arabic{enumi})}
  \item The optimal batch size $B_{opt}$ exhibits a positive correlation with the total data amount $D$.
  \item When increasing the learning rate linearly with the batch size, a larger optimal batch size can be achieved for a specified training data amount. It echos the subsequent finding in Eq. \ref{bs-lr-eq} that larger batch sizes should be paired with higher learning rates. 
\end{enumerate}

We connect the red points using linear increasing learning rate\footnote{As suggested in Sec. \ref{case4}, as the batch sizes increase, the linear learning rate is the best scheme out of the three considered. Moreover, only the red points under the linear learning rate scheme generally fall within the empirically observed batch size range rather than the predicted one, i.e., represented by solid curves instead of the dotted ones. } in Fig. \ref{law_case3_1_all},  and then establish the relationship between the optimal batch sizes 
and the available training data amount in Fig. \ref{law_fig_case3_1_2}. It reveals two main trends:

First, when the training data amount is relatively limited (e.g., less than 10B), the optimal batch size increases linearly with the data amount (i.e., the exponent term of $D$ is 1),  meaning that the number of iterations should remain above a minimum threshold. 
Also, as discussed in Sec. \ref{related_cri_bs}, there exist a minimum iteration step to achieve a certain loss. In Fig. \ref{law_case3_1_all}, this minimum step ranges approximately between 2,500 and 6,000. 
Second, as the training data amount increases, the optimal batch size grows sub-linearly with the data amount. Specifically:

\begin{equation}\label{D_BS}
B_{opt} \approx 3.24 \times 10^3 \cdot D^{0.264}
\end{equation}
\textbf{Notably, this relationship will not yield FLOP-Loss optimality. Instead, it applies to scenarios where both $N$ and $D$ are pre-determined and may not necessarily satisfy the compute-efficient frontier condition in Eq. \ref{our_L_opt_alpha}. }

Following this guideline, a batch size of approximately 4.7M is suggested for 1T training tokens, and about 8.7M for 10T training tokens.

\subsection{The Law between B and LR} \label{case4}
\begin{figure}[t]
    \centering
    \includegraphics[width=0.95\linewidth]{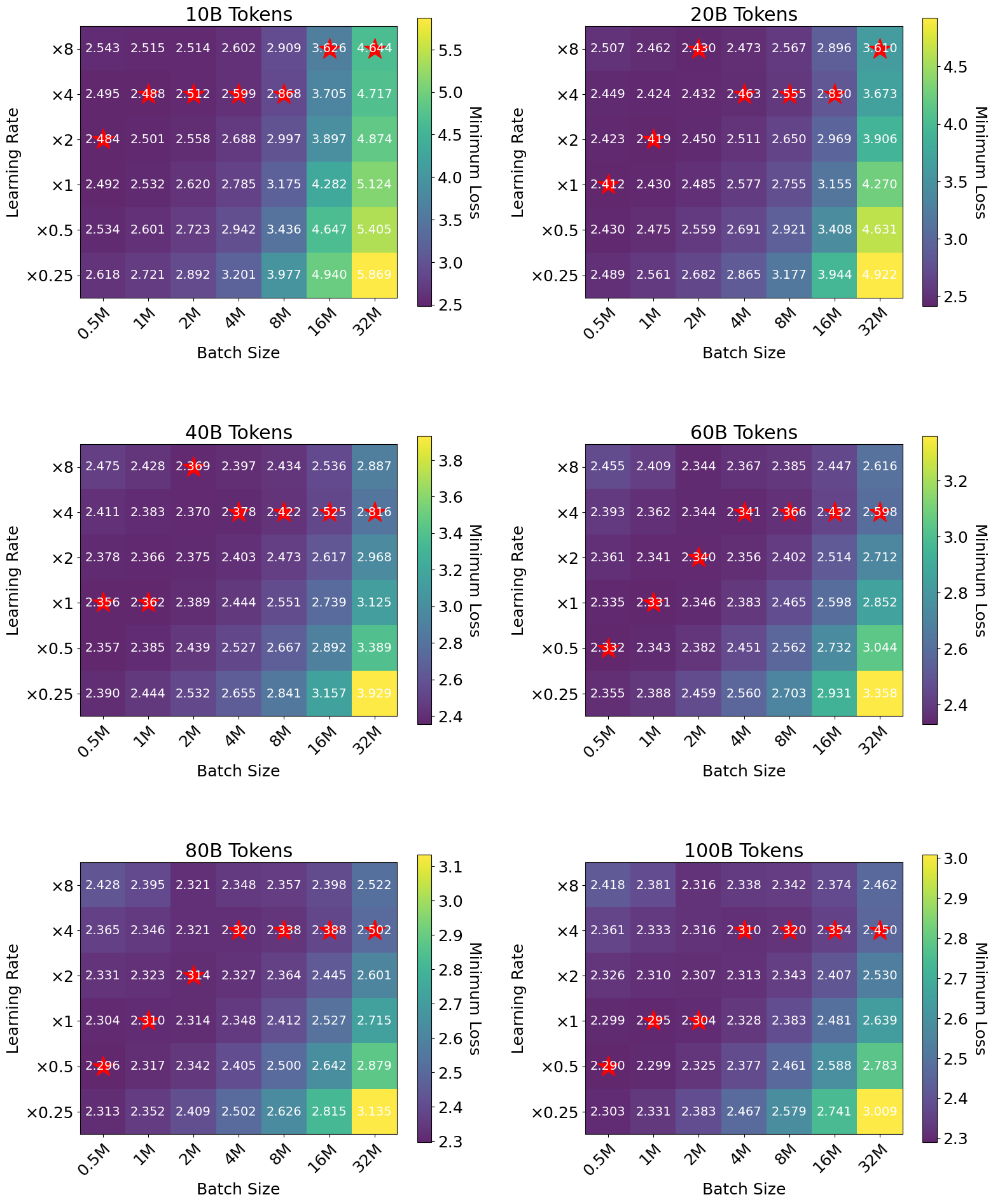}
    \caption{Training losses under different learning rates and batch sizes, with 10B to 100B training tokens. Red stars denote the lowest loss of a certain batch size. The ×1 learning rate means the corresponding one of 350M model in Table \ref{gpt3_table}.  }
    \label{law_case4_1}
\end{figure}

\begin{figure}[t]
    \centering
    \includegraphics[width=0.9\linewidth]{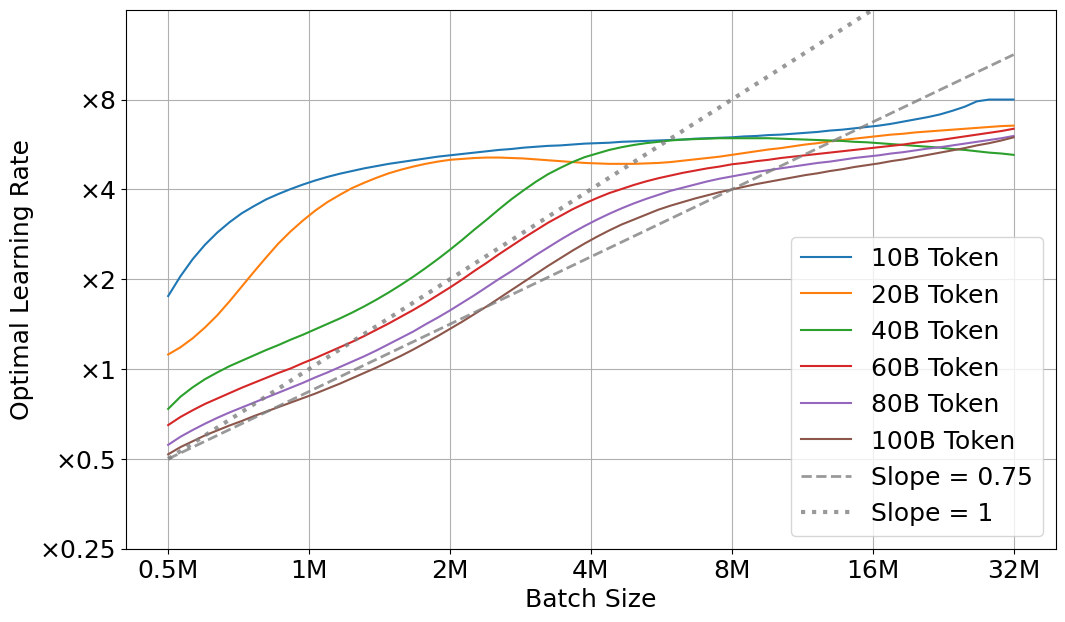}
    \caption{The optimal learning rate across different batch sizes, varying the number of tokens. }
    \label{law_case4_3}
\end{figure}

In Fig. \ref{law_case4_1}, we perform a grid search on batch size and learning rate for a 350M model trained on up to 100B tokens. The symbol “×" in y-axis means the learning rate scaling factor at every iteration throughout the training. We have two observations: 
\begin{enumerate}
\renewcommand{\labelenumi}{\arabic{enumi})}
  \item The optimal learning rate initially increases with batch size, but this growth gradually slows and eventually plateaus as batch size continues to increase. 
  \item When the training data amount is small, a higher learning rate tends to be better.
\end{enumerate}
In order to establish a continuous relationship between the learning rate and batch size, we employ two-dimensional interpolation to approximate the 3D loss surface, as illustrated in Fig. \ref{law_case4_2}. The resulting optimal learning rates are presented in Fig. \ref{law_case4_3}. 
We attribute the change in optimal learning rate along with the batch size into three primary factors:
\begin{enumerate}
\renewcommand{\labelenumi}{\arabic{enumi})}
\item \textbf{Larger batch size reduces the gradient noise scale (see Sec. \ref{grad_noise_scale}), enabling the use of higher learning rates without causing divergence.} This aligns with the finding in previous works \cite{donotdecay2018,minicpm} that increasing batch size could be an alternative to diminishing learning rate. 
\item \textbf{Given a fixed amount of training data, the use of large batch sizes leads to fewer training steps}. This reduction in steps itself requires a compensation for larger learning rate, because some training steps during the rapid training loss decline period are eliminated. This is also evidenced by the downward shift from curves of 10B to 100B tokens in Fig. \ref{law_case4_3}. Based on the derivation in Sec. \ref{case1}, we anticipate that the 10B-token curve is affected by the lack of training steps when the batch size exceeds 1M. 
\item \textbf{Even if the batch size is large enough, there exists a theoretical upper limit to the learning rate}. If the learning rate continues increasing, the model performance will degrade, as described in Eq. \ref{max_lr_eq}. This explains why the curve plateaus over large batch sizes, even if the above two factors tend to enlarge the optimal batch size.
\end{enumerate}

For the 350M model we test, when the learning rate is approaching 2.4e-3 (i.e., the scaling factor is ×8), the third factor dominates the second one. Otherwise, 
we find the optimal learning rate increases sub-linearly with the batch size:

\begin{equation} \label{bs-lr-eq}
{LR}_{opt} \propto B^{\gamma}, \quad where \ \gamma \in [0.75, 1]
\end{equation}
Although we have not yet conducted the same experiments across varying model sizes, as proposed by \cite{mutuning2021}, when using Maximal Update Parametrization, many optimal hyper-parameter including the learning rate and batch size obtained from a small model could be directly transferred to other larger model size.  
We anticipate there is a general principle for determining the appropriate learning rate for an unseen large batch size:
1) First, run a normal batch size baseline and find an optimal learning rate. 
2) Increase the learning rate sub-linearly with the new batch size, as long as it does not exceed the upper limit of the learning rate of this size of model. 

\subsection{Extrapolation Experiment} \label{case5}
We summary our findings law as follows:
\begin{enumerate}
\renewcommand{\labelenumi}{\arabic{enumi})}
\item With sufficient amount of data and appropriate model size, \textbf{to achieve compute-loss frontier} (repeat Eq. \ref{our_L_opt_alpha} and Eq. \ref{our_opt_bs}):

\begin{multline}
N_{opt} \approx 0.297 \cdot C^{0.464}, \quad D_{opt} \approx 0.561 \cdot C^{0.536}, \\
S_{opt} \approx 8.74 \times 10^{-5} \cdot C^{0.434}, \quad B_{opt} \approx 6.42 \times 10^{3} \cdot C^{0.102}
\label{case5_eq1}
\end{multline}

\item With given amount of data, \textbf{while $N$ and $D$ do not necessarily satisfy Eq. \ref {case5_eq1}} (repeat Eq. \ref{D_BS}):
\begin{equation}\label{D_BS_new}
S_{opt} \approx 3.09 \times 10^{-4} \cdot D^{0.736}, \quad B_{opt} \approx 3.24 \times 10^3 \cdot D^{0.264}
\end{equation}
\end{enumerate}

\begin{figure}[t]
    \centering
    \includegraphics[width=0.9\linewidth]{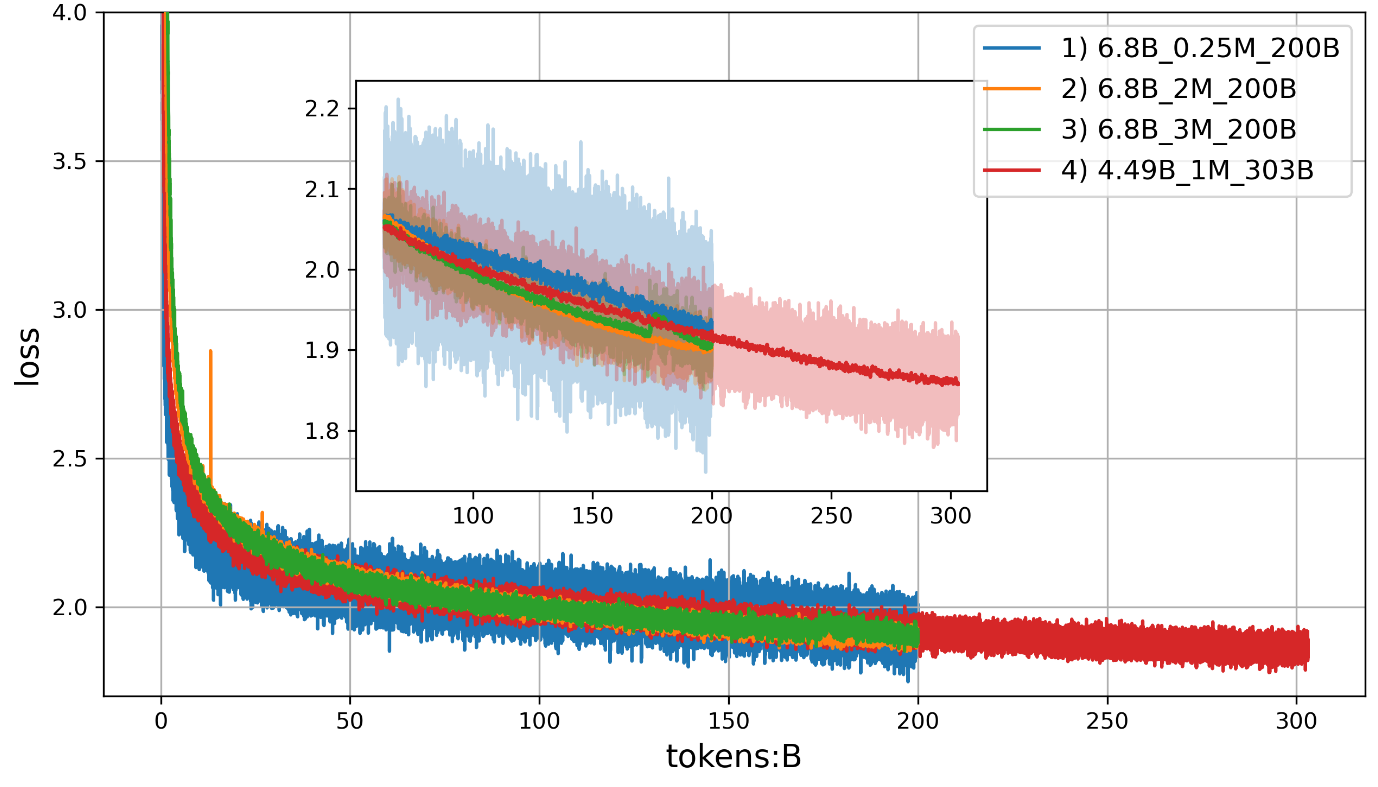}
    \caption{The training loss of three baselines, with the same total FLOPs. The 6.8B models stop at 200B tokens, while the 4.49B model stops at 303B tokens. All of them have the same total training FLOPs $8.16 \times 10^{21}$. We observe a loss spike in the third baseline near the end of the training, which potentially degrades the final performance.  }
    \label{extra_1}
\end{figure}

We then use these findings and extrapolate to larger models. The details of models are presented in Table \ref{table_extrapolate}.\footnote{Under precise calculation, baseline 3) should use a batch size of 3.12M tokens. Baseline 4) should use the model size of 4.36B parameters, 311.78B tokens of training data, and a batch size of 1.10M tokens. We use a nearby approximation in practice.} 
Figure \ref{extra_1} shows the training loss curves of these four baselines, and Table \ref{extrapolate_downstream}  present the accuracy on downstream tasks. 

We find under the same compute budget, the forth baseline achieves the lowest loss and the highest downstream task performance, which aligns with prediction in Eq. \ref{case5_eq1}. 
While using the 6.8B model and the fixed amount of 200B tokens, utilizing a suggested global batch size from Eq. \ref{bs-lr-eq} (i.e., 3M) is better than using a small batch size (i.e., 0.25M). This validates our predicted laws. 

\begin{table}[t]
\centering
\begin{tabular}{|c|c|c|c|c|c|c|}
\hline
 & $N$ & $D$ & FLOPs & $B$ & $LR$ \\
\hline
1) & 6.80B & 200B & $ 8.16 \times 10^{21}$ & 0.25M  & $1.2 \times 10^{-4}$\\
2) & 6.80B & 200B & $ 8.16 \times 10^{21}$ & 2M  & $1.2 \times 10^{-4}$ \\
3) & 6.80B & 200B & $ 8.16 \times 10^{21}$ & 3M  & $1.8 \times 10^{-4}$\\
4) & 4.49B & 303B & $ 8.16 \times 10^{21}$ & 1M  & $1.2 \times 10^{-4}$\\
\hline
\end{tabular}
\caption{The first baseline is a small batch size setting. The second one is the setting adopted by GPT-3. 
For The third baseline, the batch size is obtained using Eq. \ref{D_BS_new}, while the learning rate is adjusted according to the linear scaling rule, as suggested in Eq. \ref{bs-lr-eq}. 
The forth baseline is derived using Eq. \ref{case5_eq1}. Both 4.49B and 6.80B models have 32 attention heads and 32 layers, differing only in the hidden size: 3328 and 4096, respectively. 
}
\label{table_extrapolate}
\end{table}

\begin{table}[t] 
\centering 
\begin{tabular}{|c|c|c|c|c|} 
\hline 
& 1) 6.8B- & 2) 6.8B- & 3) 6.8B- & 4) 4.49B- \\ 
& 0.25M & 2M & 3M & 1M \\ 
\hline 
Lambada & 65.7 & 67.4 & 67.7 & 68.1 \\ 
HellaSwag & 62.7 & 62.0 & 63.3 & 65.9 \\ 
PIQA & 75.3 & 75.4 & 75.7 & 76.2 \\ 
Winograd & 63.9 & 63.4 & 62.9 & 66.7 \\ 
MMLU & 26.6 & 27.5 & 28.1 & 27.7 \\ 
\hline 
Average & 58.8 & 59.1 & 59.5 & 60.9 \\ 
\hline 
\end{tabular} 
\caption{Downstream performance of four baselines. } 
\label{extrapolate_downstream} 
\end{table}


\section{Discussion \& Conclusion}

\textbf{Discussion.}
Our scaling law research may have several limitations. 
First, we use models with sizes from 125M to 2.6B to predict the law across FLOPs from 1e18 to 1e21. However, as mentioned by \cite{chinchilla}, using different part of frontier-points could yield different envelopes. In this work, we do not take it into account. 
Also, unlike \cite{chinchilla}, we do not use several methods to estimate the optimal parameter/training tokens allocation.  In the estimation of $L(N,D)$, we leverage the internal relationship of the obtained $N_{opt}$ and $D_{opt}$, which may suffer cumulative error. 
In Sec. \ref{case3}, do not obtain a explicit relationship between the optimal batch size and the model size, while in Fig. \ref{law_fig_case3_1_2}, different size models show a subtle difference in their trends. 
Additionally, we use constant batch size throughout each training. To adjust the learning rate of every step, we use a learning rate scaling factor instead of different learning rate warm-up and decay schedules, which considered an influence of the training progress \cite{scalinglawlearningrate,minicpm,dis_scaling_law}. 
There are also other hyper-parameters we do not considered that could influence the obtained law \cite{dis_scaling_law}, such as the AdamW $\beta2$ parameter. 
Lastly, the extrapolation experiment do not involves models larger than 7B or with more training data, considering the compute resources .

\noindent
\textbf{Conclusion.}
On Huawei Ascend clusters, we train language models with parameters ranging from 125 million to 2.6 billion , using up to 300 billion high-quality tokens. These experiments enables us to establish a basic scaling laws on the optimal model size and training data amount under a given compute budget. 
We then study the impact of varying batch sizes and learning rates.
Our analysis reveals the batch size scaling laws under two cases: with fixed compute budget, the optimal batch size adheres to a power function of the compute $C$, while with fixed amount of training data, it follows a power function to data amount $D$. 
We also find that the optimal learning rate grows sub-linearly with the batch size. Finally, we use extrapolation experiments on models of increasing sizes to validate our predicted laws.


\bibliographystyle{named}
\bibliography{ijcai24}

\begin{thebibliography}{}

\bibitem[\protect\citeauthoryear{Anil \bgroup \em et al.\egroup }{2023}]{palm2}
Rohan Anil, Andrew~M. Dai, Orhan Firat, Melvin Johnson, Dmitry Lepikhin, Alexandre Passos, Siamak Shakeri, Emanuel Taropa, Paige Bailey, Zhifeng Chen, Eric Chu, Jonathan~H. Clark, Laurent~El Shafey, Yanping Huang, Kathy Meier-Hellstern, Gaurav Mishra, Erica Moreira, Mark Omernick, Kevin Robinson, Sebastian Ruder, Yi~Tay, Kefan Xiao, Yuanzhong Xu, Yujing Zhang, Gustavo~Hernandez Abrego, Junwhan Ahn, Jacob Austin, Paul Barham, Jan Botha, James Bradbury, Siddhartha Brahma, Kevin Brooks, Michele Catasta, Yong Cheng, Colin Cherry, Christopher~A. Choquette-Choo, Aakanksha Chowdhery, Clément Crepy, Shachi Dave, Mostafa Dehghani, Sunipa Dev, Jacob Devlin, Mark Díaz, Nan Du, Ethan Dyer, Vlad Feinberg, Fangxiaoyu Feng, Vlad Fienber, Markus Freitag, Xavier Garcia, Sebastian Gehrmann, Lucas Gonzalez, Guy Gur-Ari, Steven Hand, Hadi Hashemi, Le~Hou, Joshua Howland, Andrea Hu, Jeffrey Hui, Jeremy Hurwitz, Michael Isard, Abe Ittycheriah, Matthew Jagielski, Wenhao Jia, Kathleen Kenealy, Maxim Krikun, Sneha Kudugunta, Chang
  Lan, Katherine Lee, Benjamin Lee, Eric Li, Music Li, Wei Li, YaGuang Li, Jian Li, Hyeontaek Lim, Hanzhao Lin, Zhongtao Liu, Frederick Liu, Marcello Maggioni, Aroma Mahendru, Joshua Maynez, Vedant Misra, Maysam Moussalem, Zachary Nado, John Nham, Eric Ni, Andrew Nystrom, Alicia Parrish, Marie Pellat, Martin Polacek, Alex Polozov, Reiner Pope, Siyuan Qiao, Emily Reif, Bryan Richter, Parker Riley, Alex~Castro Ros, Aurko Roy, Brennan Saeta, Rajkumar Samuel, Renee Shelby, Ambrose Slone, Daniel Smilkov, David~R. So, Daniel Sohn, Simon Tokumine, Dasha Valter, Vijay Vasudevan, Kiran Vodrahalli, Xuezhi Wang, Pidong Wang, Zirui Wang, Tao Wang, John Wieting, Yuhuai Wu, Kelvin Xu, Yunhan Xu, Linting Xue, Pengcheng Yin, Jiahui Yu, Qiao Zhang, Steven Zheng, Ce~Zheng, Weikang Zhou, Denny Zhou, Slav Petrov, and Yonghui Wu.
\newblock Palm 2 technical report, 2023.

\bibitem[\protect\citeauthoryear{Brown \bgroup \em et al.\egroup }{2020}]{gpt3}
Tom Brown, Benjamin Mann, Nick Ryder, Melanie Subbiah, Jared~D Kaplan, Prafulla Dhariwal, Arvind Neelakantan, Pranav Shyam, Girish Sastry, Amanda Askell, Sandhini Agarwal, Ariel Herbert-Voss, Gretchen Krueger, Tom Henighan, Rewon Child, Aditya Ramesh, Daniel Ziegler, Jeffrey Wu, Clemens Winter, Chris Hesse, Mark Chen, Eric Sigler, Mateusz Litwin, Scott Gray, Benjamin Chess, Jack Clark, Christopher Berner, Sam McCandlish, Alec Radford, Ilya Sutskever, and Dario Amodei.
\newblock Language models are few-shot learners.
\newblock In H.~Larochelle, M.~Ranzato, R.~Hadsell, M.F. Balcan, and H.~Lin, editors, {\em Advances in Neural Information Processing Systems}, volume~33, pages 1877--1901. Curran Associates, Inc., 2020.

\bibitem[\protect\citeauthoryear{DeepSeek-AI}{2024}]{deepseek-llm}
DeepSeek-AI.
\newblock Deepseek llm: Scaling open-source language models with longtermism.
\newblock {\em arXiv preprint arXiv:2401.02954}, 2024.

\bibitem[\protect\citeauthoryear{Goyal \bgroup \em et al.\egroup }{2018}]{largebs_imagenet}
Priya Goyal, Piotr Dollár, Ross Girshick, Pieter Noordhuis, Lukasz Wesolowski, Aapo Kyrola, Andrew Tulloch, Yangqing Jia, and Kaiming He.
\newblock Accurate, large minibatch sgd: Training imagenet in 1 hour, 2018.

\bibitem[\protect\citeauthoryear{Granziol \bgroup \em et al.\egroup }{2022}]{lr_func_bs_jmlr}
Diego Granziol, Stefan Zohren, and Stephen Roberts.
\newblock Learning rates as a function of batch size: A random matrix theory approach to neural network training.
\newblock {\em Journal of Machine Learning Research}, 23(173):1--65, 2022.

\bibitem[\protect\citeauthoryear{Henighan \bgroup \em et al.\egroup }{2020}]{scalinglaw_mm}
Tom Henighan, Jared Kaplan, Mor Katz, Mark Chen, Christopher Hesse, Jacob Jackson, Heewoo Jun, Tom~B. Brown, Prafulla Dhariwal, Scott Gray, Chris Hallacy, Benjamin Mann, Alec Radford, Aditya Ramesh, Nick Ryder, Daniel~M. Ziegler, John Schulman, Dario Amodei, and Sam McCandlish.
\newblock Scaling laws for autoregressive generative modeling, 2020.

\bibitem[\protect\citeauthoryear{Hernandez \bgroup \em et al.\egroup }{2021}]{scalinglaw_transfer}
Danny Hernandez, Jared Kaplan, Tom Henighan, and Sam McCandlish.
\newblock Scaling laws for transfer, 2021.

\bibitem[\protect\citeauthoryear{Hoffer \bgroup \em et al.\egroup }{2017}]{hoffer2017train}
Elad Hoffer, Itay Hubara, and Daniel Soudry.
\newblock Train longer, generalize better: closing the generalization gap in large batch training of neural networks.
\newblock {\em Advances in neural information processing systems}, 30, 2017.

\bibitem[\protect\citeauthoryear{Hoffmann \bgroup \em et al.\egroup }{2022}]{chinchilla}
Jordan Hoffmann, Sebastian Borgeaud, Arthur Mensch, Elena Buchatskaya, Trevor Cai, Eliza Rutherford, Diego de~Las~Casas, Lisa~Anne Hendricks, Johannes Welbl, Aidan Clark, Tom Hennigan, Eric Noland, Katie Millican, George van~den Driessche, Bogdan Damoc, Aurelia Guy, Simon Osindero, Karen Simonyan, Erich Elsen, Jack~W. Rae, Oriol Vinyals, and Laurent Sifre.
\newblock Training compute-optimal large language models, 2022.

\bibitem[\protect\citeauthoryear{Hu \bgroup \em et al.\egroup }{2024}]{minicpm}
Shengding Hu, Yuge Tu, Xu~Han, Chaoqun He, Ganqu Cui, Xiang Long, Zhi Zheng, Yewei Fang, Yuxiang Huang, Weilin Zhao, Xinrong Zhang, Zheng~Leng Thai, Kaihuo Zhang, Chongyi Wang, Yuan Yao, Chenyang Zhao, Jie Zhou, Jie Cai, Zhongwu Zhai, Ning Ding, Chao Jia, Guoyang Zeng, Dahai Li, Zhiyuan Liu, and Maosong Sun.
\newblock Minicpm: Unveiling the potential of small language models with scalable training strategies, 2024.

\bibitem[\protect\citeauthoryear{Jiang \bgroup \em et al.\egroup }{2024}]{megascale}
Ziheng Jiang, Haibin Lin, Yinmin Zhong, Qi~Huang, Yangrui Chen, Zhi Zhang, Yanghua Peng, Xiang Li, Cong Xie, Shibiao Nong, Yulu Jia, Sun He, Hongmin Chen, Zhihao Bai, Qi~Hou, Shipeng Yan, Ding Zhou, Yiyao Sheng, Zhuo Jiang, Haohan Xu, Haoran Wei, Zhang Zhang, Pengfei Nie, Leqi Zou, Sida Zhao, Liang Xiang, Zherui Liu, Zhe Li, Xiaoying Jia, Jianxi Ye, Xin Jin, and Xin Liu.
\newblock {MegaScale}: Scaling large language model training to more than 10,000 {GPUs}.
\newblock In {\em 21st USENIX Symposium on Networked Systems Design and Implementation (NSDI 24)}, pages 745--760, Santa Clara, CA, April 2024. USENIX Association.

\bibitem[\protect\citeauthoryear{Kaplan \bgroup \em et al.\egroup }{2020}]{scalinglawopenai2020}
Jared Kaplan, Sam McCandlish, Tom Henighan, Tom~B Brown, Benjamin Chess, Rewon Child, Scott Gray, Alec Radford, Jeffrey Wu, and Dario Amodei.
\newblock Scaling laws for neural language models.
\newblock {\em arXiv preprint arXiv:2001.08361}, 2020.

\bibitem[\protect\citeauthoryear{Keskar \bgroup \em et al.\egroup }{2017}]{largebs_generationgap}
Nitish~Shirish Keskar, Dheevatsa Mudigere, Jorge Nocedal, Mikhail Smelyanskiy, and Ping Tak~Peter Tang.
\newblock On large-batch training for deep learning: Generalization gap and sharp minima.
\newblock In {\em International Conference on Learning Representations}, 2017.

\bibitem[\protect\citeauthoryear{Krajewski \bgroup \em et al.\egroup }{2024}]{scalinglaw_moe}
Jakub Krajewski, Jan Ludziejewski, Kamil Adamczewski, Maciej Pióro, Michał Krutul, Szymon Antoniak, Kamil Ciebiera, Krystian Król, Tomasz Odrzygóźdź, Piotr Sankowski, Marek Cygan, and Sebastian Jaszczur.
\newblock Scaling laws for fine-grained mixture of experts, 2024.

\bibitem[\protect\citeauthoryear{Lewkowycz \bgroup \em et al.\egroup }{2020}]{lewkowycz2020large}
Aitor Lewkowycz, Yasaman Bahri, Ethan Dyer, Jascha Sohl-Dickstein, and Guy Gur-Ari.
\newblock The large learning rate phase of deep learning: the catapult mechanism.
\newblock {\em arXiv preprint arXiv:2003.02218}, 2020.

\bibitem[\protect\citeauthoryear{Li \bgroup \em et al.\egroup }{2024}]{li2024surge}
Shuaipeng Li, Penghao Zhao, Hailin Zhang, Xingwu Sun, Hao Wu, Dian Jiao, Weiyan Wang, Chengjun Liu, Zheng Fang, Jinbao Xue, Yangyu Tao, Bin Cui, and Di~Wang.
\newblock Surge phenomenon in optimal learning rate and batch size scaling, 2024.

\bibitem[\protect\citeauthoryear{Lin \bgroup \em et al.\egroup }{2020}]{lin2020extrapolation}
Tao Lin, Lingjing Kong, Sebastian Stich, and Martin Jaggi.
\newblock Extrapolation for large-batch training in deep learning.
\newblock In {\em International Conference on Machine Learning}, pages 6094--6104. PMLR, 2020.

\bibitem[\protect\citeauthoryear{McCandlish \bgroup \em et al.\egroup }{2018}]{criticalbs2018}
Sam McCandlish, Jared Kaplan, Dario Amodei, and OpenAI~Dota Team.
\newblock An empirical model of large-batch training.
\newblock {\em arXiv preprint arXiv:1812.06162}, 2018.

\bibitem[\protect\citeauthoryear{{Meta AI}}{2024}]{llama3_herd}
{Meta AI}.
\newblock The llama 3 herd of models.
\newblock Technical report, Meta, 2024.

\bibitem[\protect\citeauthoryear{Porian \bgroup \em et al.\egroup }{2024a}]{dis_kaplan_chichinlla}
Tomer Porian, Mitchell Wortsman, Jenia Jitsev, Ludwig Schmidt, and Yair Carmon.
\newblock Resolving discrepancies in compute-optimal scaling of language models, 2024.

\bibitem[\protect\citeauthoryear{Porian \bgroup \em et al.\egroup }{2024b}]{dis_scaling_law}
Tomer Porian, Mitchell Wortsman, Jenia Jitsev, Ludwig Schmidt, and Yair Carmon.
\newblock Resolving discrepancies in compute-optimal scaling of language models, 2024.

\bibitem[\protect\citeauthoryear{Smith \bgroup \em et al.\egroup }{2018}]{donotdecay2018}
Sam Smith, Pieter jan Kindermans, Chris Ying, and Quoc~V. Le.
\newblock Don't decay the learning rate, increase the batch size.
\newblock 2018.

\bibitem[\protect\citeauthoryear{Smith \bgroup \em et al.\egroup }{2022}]{mtnlg}
Shaden Smith, Mostofa Patwary, Brandon Norick, Patrick LeGresley, Samyam Rajbhandari, Jared Casper, Zhun Liu, Shrimai Prabhumoye, George Zerveas, Vijay Korthikanti, Elton Zhang, Rewon Child, Reza~Yazdani Aminabadi, Julie Bernauer, Xia Song, Mohammad Shoeybi, Yuxiong He, Michael Houston, Saurabh Tiwary, and Bryan Catanzaro.
\newblock Using deepspeed and megatron to train megatron-turing nlg 530b, a large-scale generative language model, 2022.

\bibitem[\protect\citeauthoryear{Tissue \bgroup \em et al.\egroup }{2024}]{scalinglawlearningrate}
Howe Tissue, Venus Wang, and Lu~Wang.
\newblock Scaling law with learning rate annealing, 2024.

\bibitem[\protect\citeauthoryear{Wen \bgroup \em et al.\egroup }{2018}]{wen2018smoothout}
Wei Wen, Yandan Wang, Feng Yan, Cong Xu, Chunpeng Wu, Yiran Chen, and Hai Li.
\newblock Smoothout: Smoothing out sharp minima to improve generalization in deep learning, 2018.

\bibitem[\protect\citeauthoryear{Yang \bgroup \em et al.\egroup }{2021}]{mutuning2021}
Ge~Yang, Edward Hu, Igor Babuschkin, Szymon Sidor, Xiaodong Liu, David Farhi, Nick Ryder, Jakub Pachocki, Weizhu Chen, and Jianfeng Gao.
\newblock Tuning large neural networks via zero-shot hyperparameter transfer.
\newblock {\em Advances in Neural Information Processing Systems}, 34:17084--17097, 2021.

\end{thebibliography}

\appendix
\section{Approach for Obtaining Eq. \ref{our_L_ND}}\label{appen_a}
\subsection{Internal relationship between the coefficients}
Given $L(N, D) = E + \frac{A}{N^{\alpha}} + \frac{B}{D^{\beta}}$, achieving the optimal $L$ requires $\frac{\partial{L}}{\partial{N}}|_{N=N_{opt}}=0$. After combining it with the following equations:
\begin{itemize}
  \item $\frac{\partial{L(N,D)}}{\partial{N}}= -A\alpha N^{-(\alpha+1)}-B\beta D^{-(\beta+1)} \frac{\partial{D}}{\partial{N}}$
  \item For any fixed $C$, $\frac{\partial{D}}{\partial{N}} = \frac{\partial{(C/(6N))}}{\partial{N}}=-\frac{D}{N}$
  \item Given the compute $C$, to achieve the optimal $L$, we have $N_{opt} = p \cdot C^{a}, D_{opt} = q \cdot C^{b}$, where $pq=6, a+b=1$
\end{itemize}

\noindent
we can have 
$$
A\alpha q^{\beta} C^{\beta b} = B \beta p^{\alpha} C^{\alpha a}, \quad \forall C
$$
To let it hold for any $C$, we have:
\begin{equation} \label{huber_condition}
\frac{\alpha}{\beta} = \frac{b}{a}, \quad \frac{A}{B}=\frac{\beta p^{\alpha}}{\alpha q^{\beta}}
\end{equation}

\subsection{Minimizing the Huber Loss}
Following \cite{chinchilla}, we employ the L-BFGS to estimate the parameters  $(A,B,E,\alpha,\beta)$ algorithm by minimizing the log huber loss:
$$
\min_{A,B,E,\alpha,\beta} \sum_{(N_i,D_i,L_i)} Huber_{\delta}(log L(N_i,D_i) - log L_i)
$$
We maintain the same $\delta=10^{-3}$ as adopted in \cite{chinchilla}. 
We further apply the condition derived in Eq. \ref{huber_condition} as a regularization
$$
\alpha/\beta = b/a = \frac{0.536}{0.464}, \quad log(A) = log(B) + log(\frac{\beta p^{\alpha}}{\alpha q^{\beta}})
$$
This simplifies the optimization problem to the one only involving three parameters: $(B, E, \beta)$.

In addition, we observe that the initial guess of parameters greatly affects the final regressed results. Therefore, we implement a grid search for the initialization value, and calculate the R-squared loss between the regressed loss and the true loss. 
In our observation, multiple distinct initializations with the highest R-squared values produce similar regression results, which enhances the confidence on the final results. 


\section{Hyper-Parameters mentioned in Sec. \ref{case1}}\label{appen_c}
In all experiments, we use a fixed sequence length of 2048, linear learning rate warm-up, , cosine learning rate decay, and Adam as the optimizer with $\beta_1=0.9$ and $\beta_2=0.95$. We present other configurations in Table \ref{gpt3_table}. The minimum learning rate is 1/10 of the maximum one.

\begin{table}[h]
\centering
\caption{} \label{gpt3_table}
\begin{tabular}{|c|c|c|c|}
\hline
Model & GBS & Max LR & (warmup, decay) \\
\hline
125M & 0.5M & $6.0 \times 10^{-4}$ & (715, 500,000)  \\
350M & 0.5M & $3.0 \times 10^{-4}$ & (715, 500,000) \\
760M & 0.5M & $2.5 \times 10^{-4}$ & (715, 500,000) \\
1.3B & 1M & $2.0 \times 10^{-4}$ & (350, 300,000)\\
2.6B & 1M & $1.6 \times 10^{-4}$ &  (350, 300,000)\\
\hline
\end{tabular}
\end{table}

\section{Supplemental Figures}

\begin{figure*}[t]
    \centering
    
    \begin{subfigure}[b]{\linewidth}
        \centering
        \includegraphics[width=\linewidth]{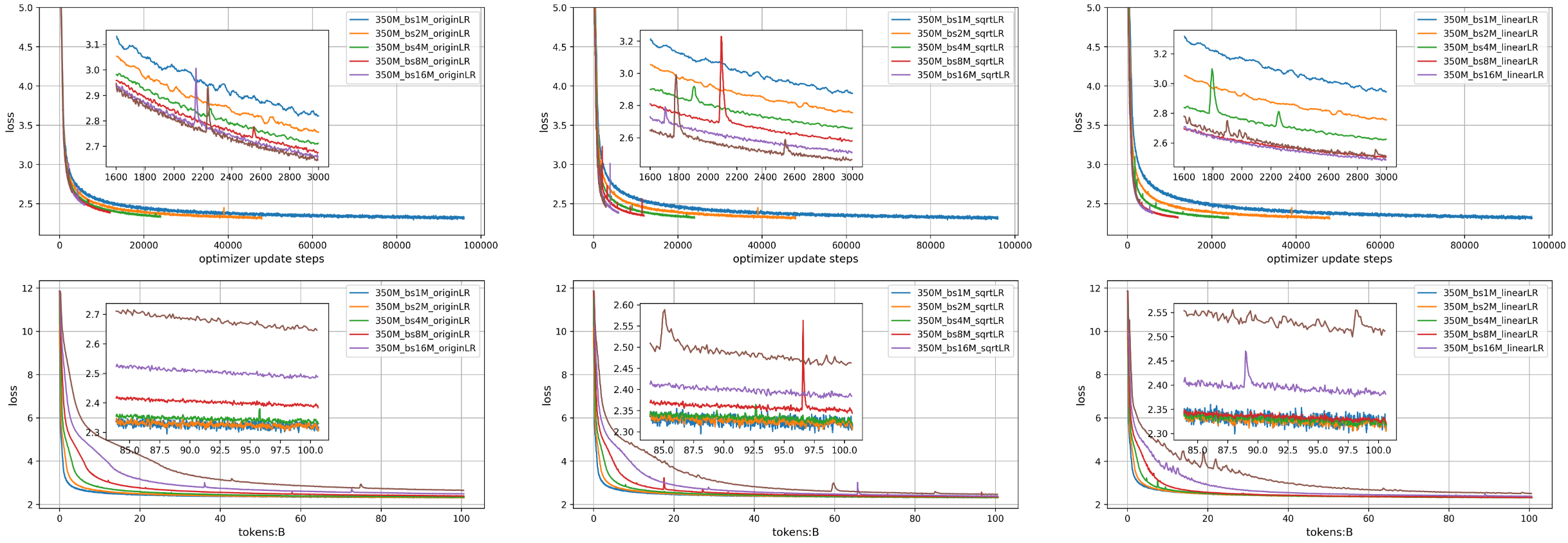}
        \caption{350M model.}
        \label{350M_main}
    \end{subfigure}
    
    \vspace{1em}
    
    \begin{subfigure}[b]{\linewidth}
        \centering
        \includegraphics[width=\linewidth]{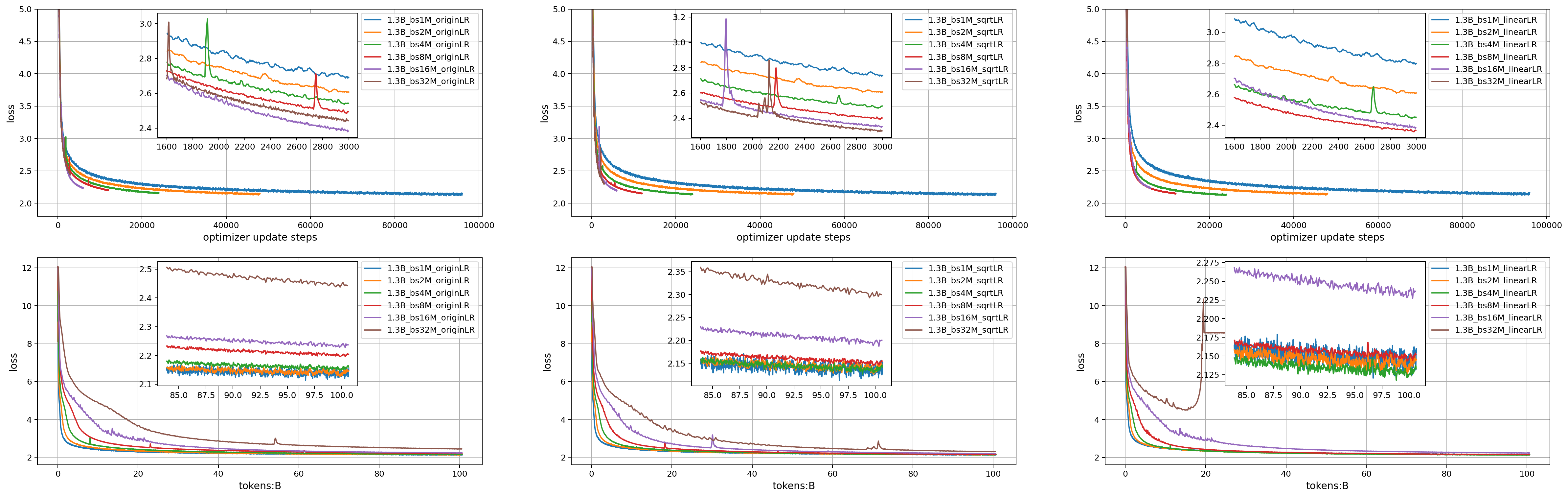}
        \caption{1.3B model.}
        \label{1.3B_main}
    \end{subfigure}
    
    \vspace{1em}
    
    \begin{subfigure}[b]{\linewidth}
        \centering
        \includegraphics[width=\linewidth]{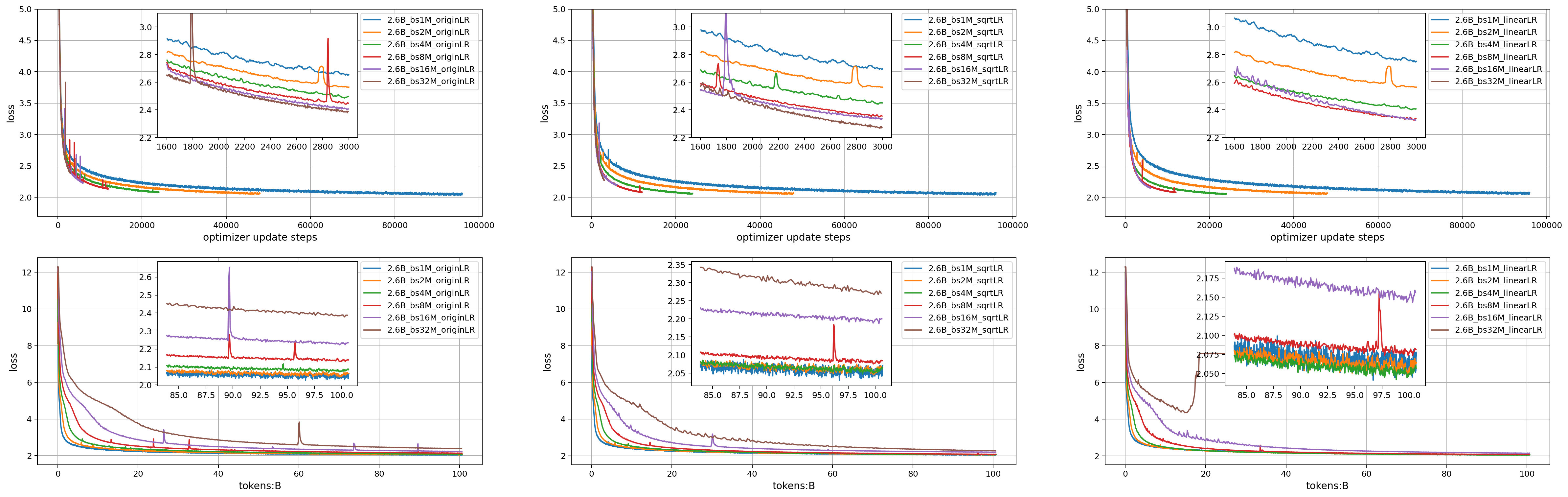}
        \caption{2.6B model.}
        \label{2.6B_main}
    \end{subfigure}
    
    \caption{Step-loss and token-loss plots for 350M , 1.3B and 2.6B models under three learning rate schemes. “OriginLR" refers to the baseline learning rate in Table \ref{gpt3_table}, while “sqrtLR" and “LinearLR" denote increasing the learning rate with the batch size in a square root manner and linear manner, respectively.  }
    \label{fig:comparison}
\end{figure*}

\subsection{Fig. \ref{fig:comparison}: More step-loss and token-loss results}\label{appen_b}

From Fig. \ref{fig:comparison}, we have several observations:
\begin{enumerate}
\renewcommand{\labelenumi}{\arabic{enumi})}
  \item Large batch sizes typically achieve lower loss compared to smaller batch sizes when considering the same optimizer update steps, while resulting in higher loss under the same number of training tokens. At 100B training tokens, a batch size larger than 8M will result in insufficient training steps. 
  We anticipate with unlimited amount of data, using large batch size for training is feasible. 
  \item Scaling the learning rate with batch size (either using square root or linear scaling) proves beneficial for large batch training. For example, in the step-loss comparison for the 1.3B model, the 32M batch size underperforms the 16M batch size with the “originLR” scheme, while outperforming it when using the “sqrtLR”. 
  In addition, increasing the learning rate diminishes the advantage of small batches (e.g. 1M global batch size) at the 100B token. For instance, in Fig. \ref{1.3B_main}, with the original learning rate, the 4M curve lies above the 1M curve, while with linear learning rate scaling, the 4M curve is below the 1M one.
  \item  Empirically, there is a limit for increasing the learning rate, as it will cause divergence issues (see the “2.6B\_bs32M\_linearLR" curve Fig. \ref{2.6B_main}). The detailed relationship between batch size and learning rate is in Sec. \ref{case4}.
\end{enumerate}

\begin{figure*}[t]
    \centering
    
    \begin{subfigure}[b]{\linewidth}
        \centering
        \includegraphics[width=\linewidth]{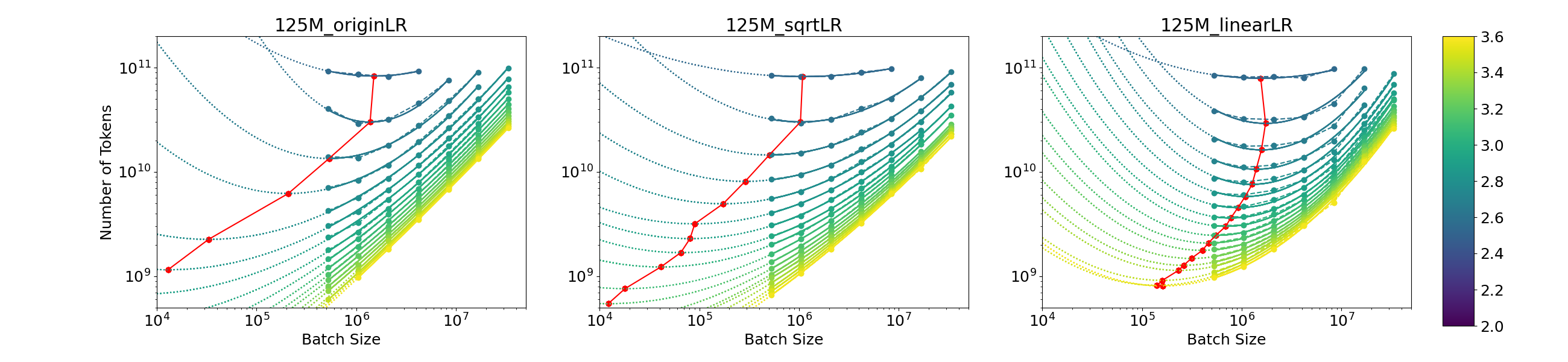}
        \caption{125M model.}
        \label{125M_law_case3_1}
    \end{subfigure}
    
    \vspace{1em}
    
    \begin{subfigure}[b]{\linewidth}
        \centering
        \includegraphics[width=\linewidth]{./pic/law_case3_1_350M.png}
        \caption{350M model.}
        \label{350M_law_case3_1}
    \end{subfigure}
    
    \vspace{1em}
    
    \begin{subfigure}[b]{\linewidth}
        \centering
        \includegraphics[width=\linewidth]{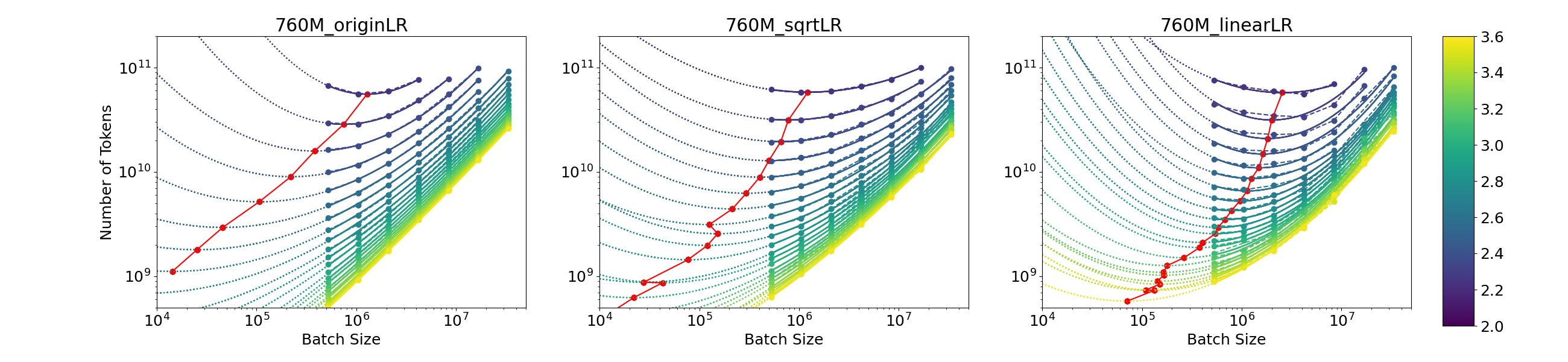}
        \caption{760M model.}
        \label{760M_law_case3_1}
    \end{subfigure}
    
    \begin{subfigure}[b]{\linewidth}
        \centering
        \includegraphics[width=\linewidth]{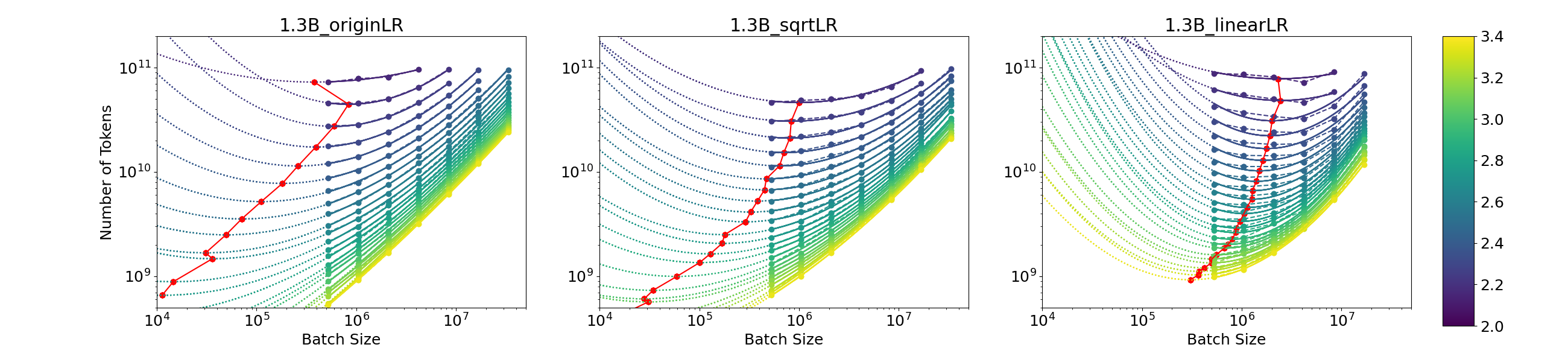}
        \caption{1.3B model.}
        \label{1_3B_law_case3_1}
    \end{subfigure}
    
    \begin{subfigure}[b]{\linewidth}
        \centering
        \includegraphics[width=\linewidth]{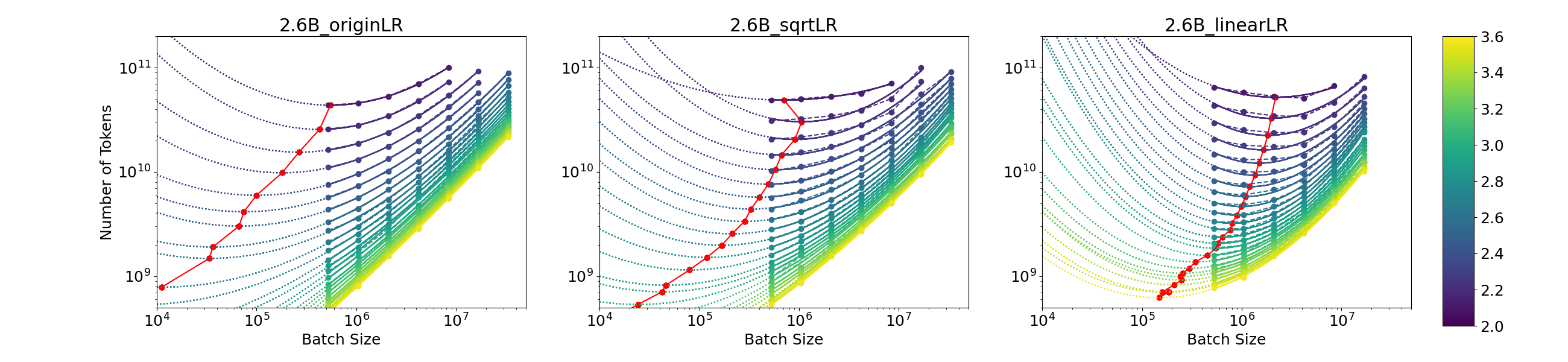}
        \caption{2.6B model.}
        \label{2_6B_law_case3_1}
    \end{subfigure}
    \caption{Loss contours of five model sizes from 125M to 2.6B size.  }
    \label{law_case3_1_all}
\end{figure*}

\subsection{Fig. \ref{law_case3_1_all}: Loss contours of five model sizes from 125M to 2.6B.}

\begin{figure*}[h]
    \centering
    \includegraphics[width=\linewidth]{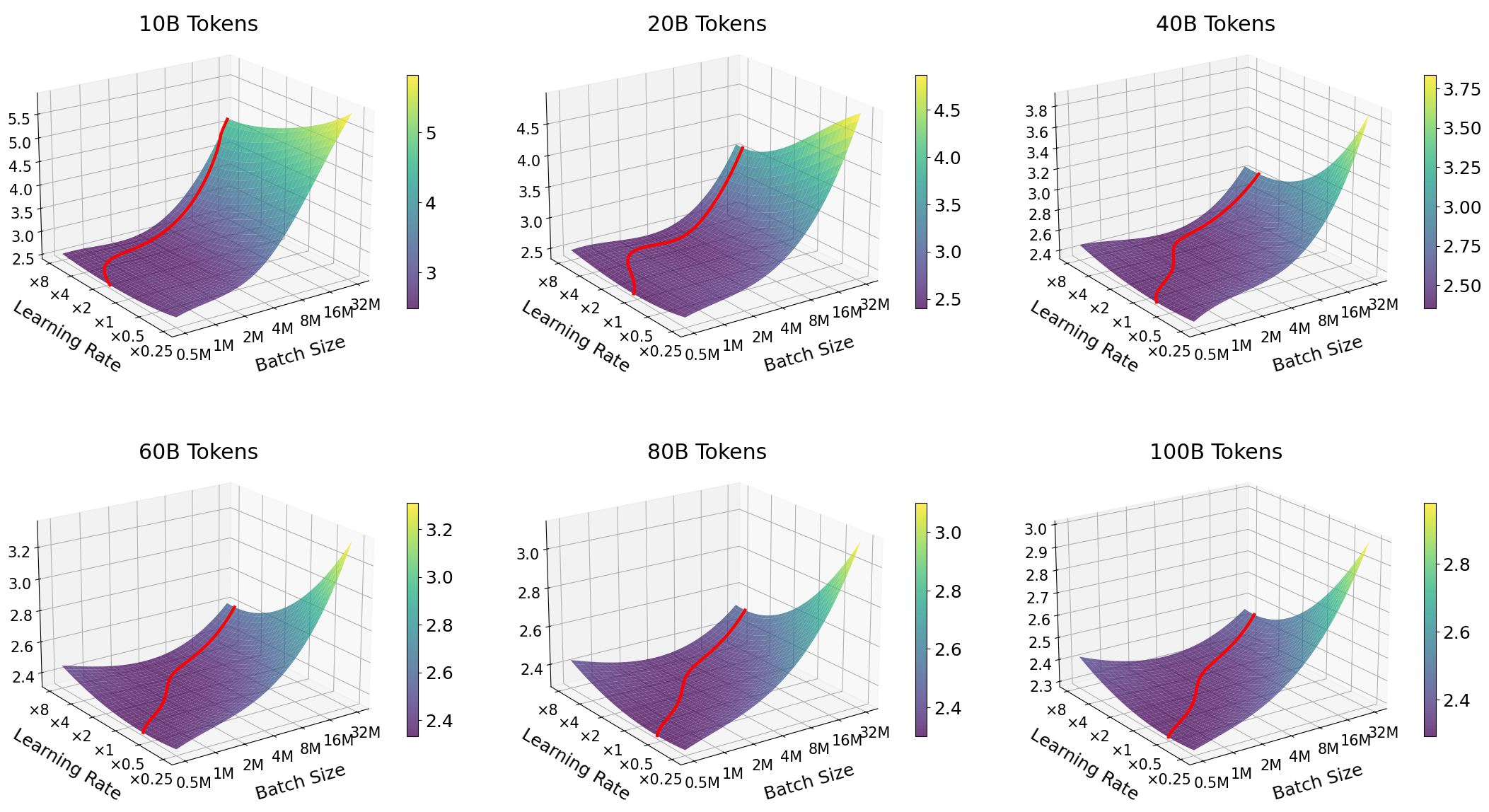}
    \caption{Loss surfaces under various combinations of batch sizes and learning rates. The red curves represent the optimal learning rate under different batch sizes, which is a continuous representation of the red stars in Fig. \ref{law_case4_1}.}
    \label{law_case4_2}
\end{figure*}

\subsection{Fig. \ref{law_case4_2}: 3D loss surface of 350M model under various combinations of batch sizes and learning rates.}

\end{document}